\documentclass{article}

\usepackage[square,numbers]{natbib}
\usepackage[final]{neurips_2025}

\usepackage{pifont}
\usepackage[utf8]{inputenc} 
\usepackage[T1]{fontenc}    
\usepackage{hyperref}       
\usepackage{url}            
\usepackage{booktabs}       
\usepackage{amsfonts}       
\usepackage{nicefrac}       
\usepackage{wrapfig}
\usepackage{microtype}      
\usepackage{xcolor}         
\usepackage[normalem]{ulem}   
\usepackage{amsmath, amssymb, amsfonts}
\usepackage{subcaption}
\usepackage{enumitem}
\usepackage[most]{tcolorbox}
\usepackage{listings}
\usepackage{pdfpages}
\usepackage{multirow}

\usepackage[toc,page,header]{appendix}
\usepackage{etoc}

\usepackage{float}

\newcommand{\ie}{\textit{i}.\textit{e}., }
\newcommand{\eg}{\textit{e}.\textit{g}., } 
\newcommand{\mmpb}{\textsc{MMPB}} 
\newcommand{\sks}{\texttt{<sks>}}

\newcommand{\concepts}{humans, animals, objects, and characters}

\usepackage{graphicx}

\title{\mmpb{}: It's Time for Multi-Modal Personalization}

\author{
Jaeik Kim$^1$ \quad Woojin Kim$^2$ \quad Woohyeon Park$^2$ \quad Jaeyoung Do$^{1,2}$ \\
AIDAS Laboratory, $^1$IPAI \& $^2$ECE, Seoul National University
\\
{\tt\small \{jake630, wjk9904, woohyeon, jaeyoung.do\}@snu.ac.kr}
}

\begin{document}

\maketitle

\begin{abstract}
Visual personalization is essential in user-facing AI systems such as smart homes and healthcare, where aligning model behavior with user-centric concepts is critical. However, recent large Vision-Language Models (VLMs), despite their broad applicability, remain underexplored in their ability to adapt to individual users. In this paper, we introduce \textbf{\mmpb{}}, the first extensive benchmark for evaluating VLMs on personalization. \mmpb{} comprises 10k image-query pairs and includes 111 personalizable concepts across four categories: \concepts{}, with the human category enriched with preference-grounded queries. We structure personalization into three main task types, each highlighting a different key property of VLMs. Using 23 widely used VLMs including both open- and closed-source models, we evaluate personalization performance via a three-stage protocol: concept injection, multi-turn dialogue, and personalized querying. Our findings indicate that most VLMs (including some closed-source models) struggle with personalization, particularly in maintaining consistency over dialogue, handling user preferences, and adapting to visual cues. Our analysis reveals that the challenges in VLM personalization (such as refusal behaviors and long-context forgetting) highlight substantial room for improvement. By identifying these limitations and offering a scalable benchmark, \mmpb{} offers valuable insights and a solid foundation for future research toward truly personalized multi-modal AI.\\
\textbf{Project Page:} \href{https://aidaslab.github.io/MMPB}{aidaslab.github.io/MMPB}

\end{abstract}

\section{Introduction}


\paragraph{Now we need VLM personalization
} Our daily lives are filled with recurring visual concepts that are closely tied to us.  
Every day, you might spot your dog waiting by the door or notice your favorite coffee mug on the table. However, they are rarely understood in a user-specific context by recent large Vision-Language Models (VLMs). Although VLMs have been widely adopted as assistants that utilize broad knowledge of the world~\cite{xu2024pllava,zhu2025internvl3,zhang2024vision,li2025benchmark}, they still operate largely under the \textit{one-size-fits-all} paradigm~\cite{achiam2023gpt,team2024gemini1.5}, treating all users as interchangeable and responding to visual inputs without adapting to individual identities, preferences, or histories. As VLMs are integrated into multi-modal systems like smart home devices and robots~\cite{fan2024exploring,kim2024openvla,zhen20243d}, they increasingly serve as the core interface for grounding user instructions in the visual world. To work effectively in everyday scenarios, they must recognize and align with each user’s visually repetitive environment and preferences. Such personalization is essential in applications where consistent interpretation of personal context is critical~\cite{nguyen2024yo,alaluf2024myvlm,an2024mc}. For example, instead of responding generically to \textit{``feed the gray tabby cat,''} the model should understand personalized commands like \textit{``feed Mochi.’’} 
Similarly, it should interpret a photo as \textit{``your favorite travel destination''} rather than \textit{``a snow-covered mountain.’’}


\paragraph{The evaluation gap in VLM personalization} Despite the growing importance of AI personalization in many real-world scenarios~\cite{duan2024manipulate,cao2024maplm,duan2024cityllava}, current benchmarks fall short in evaluating the personalization capabilities of VLMs. Existing VQA datasets focus on general-purpose knowledge, including commonsense~\cite{yu2023mm,liu2024mmbench,chen2024mmstar}, scientific~\cite{hiippala2021ai2d,lu2023mathvista,lu2022learn}, and medical reasoning~\cite{ben2019vqamed,liu2021slake,lau2018rad}. Despite some early progress~\cite{nguyen2024yo,alaluf2024myvlm}, personalization in VLMs remains constrained, with limited coverage and diversity of personalizable concepts (Table~\ref{tab:datasets}). In addition, the absence of a unified evaluation framework (both in task types and metrics) and overlooking the cold-start nature, where models must personalize with minimal prior context, make it difficult to systematically assess personalization.
Finally, preference-grounded VQA, which requires reasoning over user-specific likes and dislikes, remains especially underexplored, posing unique challenges beyond standard VLM tasks.

To bridge this gap, we introduce \textbf{\mmpb{}} (Multi-modal Personalization Benchmark), the first benchmark for evaluating VLM personalization. \mmpb{} evaluates concept recognition and preference-grounded reasoning using images associated with preference-related keywords (Figure~\ref{fig:fig1}). Our dataset is built via a human–model collaborative process guided by annotation protocols, which specify task types and structured query processing. To ensure high quality, we remove any queries solvable using only concept information or only query images, thus focusing evaluation on genuine cross-modal personalized reasoning. Consequently, \mmpb{} comprises 111 concepts spanning four categories: \concepts{}, paired with five reference images and four level textual descriptions. For the human preferences, we curate from 30 diverse subdomains of personal preferences. Queries are categorized into three task types, enabling detailed analysis of personalization failures in VLMs. Finally, we include multi-turn dialogues to evaluate the model’s ability to retain personalized concepts over extended interactions.

We evaluate 23 widely used VLMs on \mmpb{}, including closed-source models such as GPT-4o~\cite{achiam2023gpt} and Claude-Sonnet~\cite{bai2022constitutional}. Despite their strong performance on general-purpose VQA tasks, these VLMs exhibit significant limitations and challenges in personalization. Our key take-home messages include:
\textbf{(i)} Even top-ranked VLMs on general benchmarks struggle with preference-grounded tasks, underscoring the need for more robust forms of inference, including abductive reasoning
(\S\ref{subsubsec:preference_grounded_vqa}).
\textbf{(ii)} VLMs struggle with visual personalization. For example, comparable performance from one image and three text keywords highlights a persistent inability to leverage fine-grained visual cues despite the central role of image-based personalization (\S\ref{subsubsec:suboptimal_visual_concept_injection}). 
\textbf{(iii)} Closed-source models tend to avoid personalization involving human-centric concepts, stemming from safety alignment constraints. This implies that existing safety constraints hinder personalization, highlighting the need to reconsider the balance between safety and personalization (\S\ref{subsubsec:resistance_closed_models}). 
We further conduct a systematic analysis of VLM failure patterns, focusing on personalization bias and failures in discrimination (\S\ref{subsubsec:systematic_error_patterns}), with errors amplified in long-context scenarios where mid-sequence concepts are frequently overlooked (\S\ref{subsubsec:multi_turn_degradation}).
Our main contributions are as follows:

\begin{figure}
    \centering
    \includegraphics[width=\linewidth]{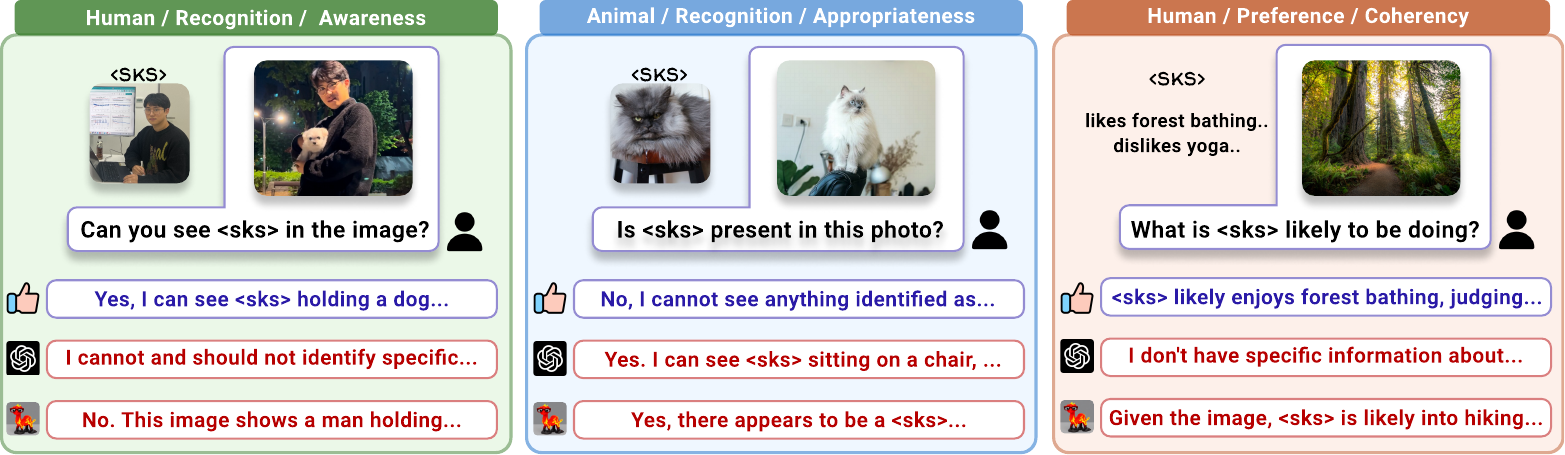}
    \caption{Examples of personalized queries across task types and representative failure cases of recent VLMs. \includegraphics[height=2ex]{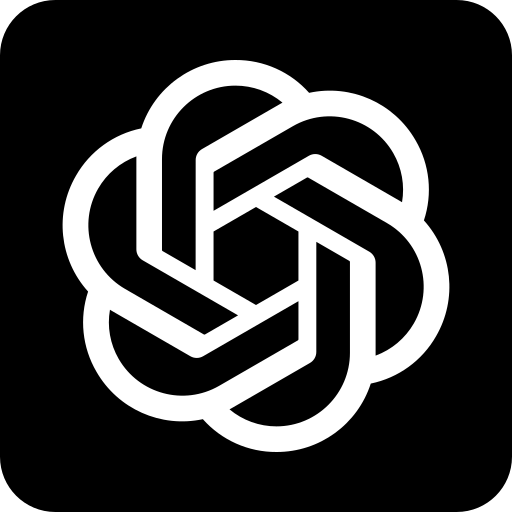} indicates GPT-4o, while \includegraphics[height=2ex]{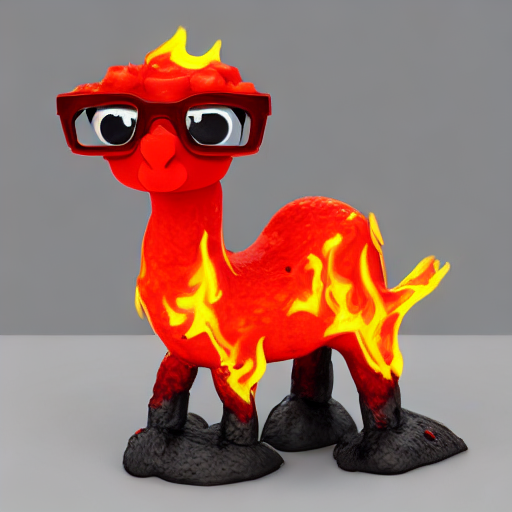} indicates LLaVA family models such as LLaVA-NeXT.}
    \label{fig:fig1}
        \vspace{-1em}
\end{figure}

\begin{itemize}

    \item We introduce \mmpb{}, the first comprehensive benchmark for evaluating VLM personalization in both recognition and preference-grounded VQA. It includes 111 personalizable concepts and 10,000+ questions with 15 task types that reflect real-world scenarios.
    \item By providing five reference images and four levels of textual descriptions, \mmpb{} establishes a foundation not only for evaluation but also for future advanced techniques that leverage them (\eg post hoc training) thereby enabling fine-grained VLM personalization.    
    \item By rigorously evaluating widely adopted VLMs, we identify key personalization failures such as limited preference-grounded reasoning, modality gaps, personalization bias, and safety-induced evasiveness, which establish \mmpb{} as a key diagnostic tool for improving personalized VLM behavior.    
\end{itemize}

\section{Problem Definition}
\label{sec:sec.2}
\subsection{Core Properties of Personalized VLMs}

To systematically evaluate personalized VLMs, we formalize four core properties we call \emph{personalization criteria}. Personalizable concepts are user-centric entities (\eg \textit{“me,”} \textit{“my dog”}) whose attributes, such as appearance or preferences, can be encoded in either structured or unstructured form. Formally, let \(\mathcal{P} = \{\,p_1, \dots, p_K\}\) denote all personalizable concepts. A VLM successfully augmented with any \(p_k \in \mathcal{P}\) should satisfy the following criteria, with examples in Figure~\ref{fig:fig1}:
\begin{enumerate}
  \item \textbf{Awareness}: Can the model accurately identify \(p_k\) in a given image?
  \item \textbf{Appropriateness}: Does the model activate \(p_k\) only when it is contextually appropriate?
  \item \textbf{Coherency}: Does the model contradict \(p_k\) in its responses?
  \item \textbf{Persistency}: Can the model consistently respond based on \(p_k\) across long-context or multi-turn interactions?
\end{enumerate}


\subsection{Formalizing VLM Personalization}
\label{sub:2.2}

A standard VLM \(\mathcal{M}\) is typically defined as a function \(f : \mathcal{X} \times \mathcal{T} \to \mathcal{Y}\), where \(\mathcal{X}\) is the pixel space, \(\mathcal{T}\) the text space (\eg user queries), and \(\mathcal{Y}\) the output space (\eg generated text). To personalize \(\mathcal{M}\), we extend it to a function \(f_p : \mathcal{X} \times \mathcal{T} \times \mathcal{P} \to \mathcal{Y}\), incorporating a concept \(p_k \in \mathcal{P}\).
A key objective of personalization is to tailor outputs to user-specific contexts while \textbf{preserving the general-purpose knowledge} acquired during pretraining. One approach is to concatenate \(p_k\) with the input modalities \(x \in \mathcal{X}\) and \(t \in \mathcal{T}\), without modifying the model’s weights. Given a projection function \(h\) that embeds \(p_k\), and a decoder \(g\), the personalized output is:
\[
y = f_{p}(x, t, p_k) = g\bigl(E_I(x) \oplus E_T(t) \oplus h(p_k)\bigr),
\]
where \(E_I\) and \(E_T\) denote the image and text encoders, with \(\oplus\) as concatenation. \mmpb{} evaluates VLMs by injecting \(p_k\) via either textual or visual modalities, using reference images or descriptions as in-context prompts~\cite{dong2022survey,hao2024remember}, using\(h = E_I\) for visual and \(h=E_T\) for textual concept injection.


\begin{table}[t]
\centering
\caption{Comparison of existing multimodal personalization benchmark datasets, indicating whether each dataset supports systemic evaluation (\emph{SysEval}), human preference (\emph{Pref}), multiple levels of granularity (\emph{Gran}), and multi-turn interactions (\emph{Multi-turn}).
\emph{\#Concept} and \emph{\#Samples} denote the number of distinct concept categories and total samples, respectively.}
\small
\setlength{\tabcolsep}{12pt}
\begin{tabular}{lcccccc}
\toprule
Dataset & SysEval & Pref & Gran & Multi-turn & \#Concept & \#Samples \\
\midrule
MyVLM~\cite{alaluf2024myvlm}    & \ding{55} & \ding{55} & \ding{55} & \ding{55} & 29  & 0.3K \\
Yo’LLaVA~\cite{nguyen2024yo} & \ding{55} & \ding{55} & \ding{55} & \ding{55} & 40  & 0.6K \\
MC-LLaVA~\cite{an2024mc} & \ding{55} & \ding{55} & \ding{55} & \ding{55} & 95  & 2.0K \\
MMPB (Ours)          & \ding{51} & \ding{51} & \ding{51} & \ding{51} & 111 & 10.0K \\
\bottomrule
\end{tabular}
\label{tab:datasets}
\end{table}

\section{\mmpb{}: Multi-modal Personalization Benchmark}
\label{sec:construct}
\subsection{Overview}

\begin{figure}
    \centering
    \includegraphics[width=\linewidth]{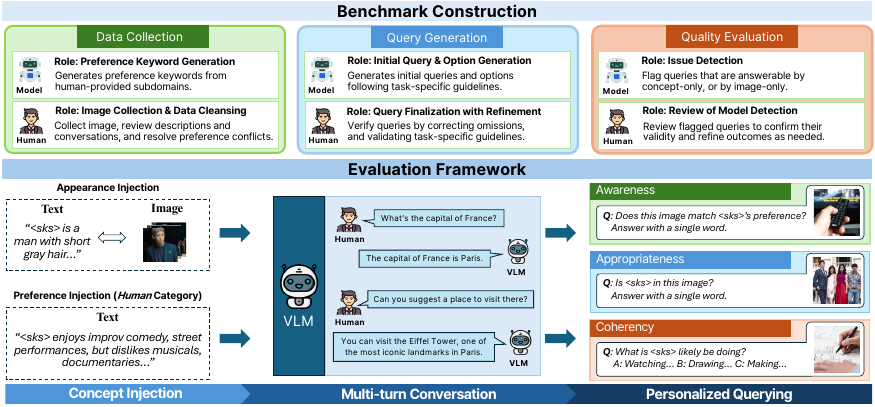}
    \caption{Overview of \mmpb{}.
    (top) A three‐step construction process ensuring high quality and scalability.
    (bottom) An evaluation protocol for assessing the VLM’s \textit{personalization criteria}.}
    \label{fig:construction}
    \vspace{-0.5em}
\end{figure}

\mmpb{} is the first benchmark for personalizing multi-modal assistant, featuring multiple-choice visual question answering (VQA) tasks focused on recognition and user preference. It comprises 10,017 image–query pairs across 111 concepts spanning four categories, covering three task types: \textit{Awareness}, \textit{Appropriateness}, and \textit{Coherency}. To support future expansion, \mmpb{} is built via a three-step human–model collaboration (Figure~\ref{fig:construction}). \mmpb{} evaluates VLMs across three stages: (1) \textbf{Concept injection}, (2) \textbf{Multi-turn conversations}, and (3) \textbf{Personalized querying}. In the first stage, a concept is introduced via reference images or textual descriptions. During the conversation phase, the model engages in general multi-turn dialogue to test concept retention. Finally, personalized queries assess whether the model can apply the concept to a visual input. Detailed dataset construction guidelines and statistics are provided in Appendix \S\ref{sup:annotation} and \S\ref{sup:statistics}.


\subsection{Data Collection}
\label{subsec:data_collection}


\paragraph{Concepts} \mmpb{} provides five reference images per concept. Human annotators, following detailed guidelines (Appendix \S\ref{sup:guide_concept_injection}), sourced these images from open datasets (MC-LLaVA~\cite{an2024mc}, MyVLM~\cite{alaluf2024myvlm}) and CC-licensed platforms (Flickr, Pexels), and recorded original URLs for any externally retrieved images (\eg kmDB or Google Image Search). \mmpb{} assumes static concept states without modeling temporal variation such as appearance changes or evolving user preferences. For consistency, most human concepts use MC‐LLaVA~\cite{an2024mc} movie portraits, ensuring a uniform character style. Other concepts use multiple images of the same instance (\eg a pink-hat Asian doll).

To facilitate concept injection through the textual modality, we generate appearance descriptions for each concept via human–model collaboration, using five reference images per concept. We employ Ovis2-34B~\cite{lu2024ovis} for its strong multi-image captioning capabilities~\cite{cheng2025caparena} and practical scalability, as it can be run efficiently on a single 80GB GPU. The model-generated descriptions are then verified and refined by human annotators based on guidelines (Appendix \S\ref{sup:text_description}). Each concept is described across four levels of granularity:  
(i) \textit{Simple} (three keywords),  
(ii) \textit{Moderate} (one sentence),  
(iii) \textit{Detailed} (one paragraph), and  
(iv) \textit{Extended} (structured multi-paragraphs). To avoid pretraining leakage and ensure fair evaluation, all concept names are replaced with \sks{}, following prior works~\cite{nguyen2024yo,ruiz2023dreambooth}.

For human preferences, we define five domains, each comprising six subdomains (30 in total; see Figure~\ref{fig:dataset_statistics} in Appendix \S\ref{sup:statistics}). Using GPT-4o~\cite{achiam2023gpt}, we extract 30–40 representative keywords per domain. Since providing multi-image preferences for every domain incurs substantial input overhead, we instead express them textually. Each human is assigned \textit{likes} and \textit{dislikes} via a template: “\sks{} likes \texttt{\{keywords\}} but dislikes \texttt{\{keywords\}}.” Conflicting preferences (\eg liking “gaming keyboards” but disliking “technical gadgets”) are resolved by human annotators.

\paragraph{Multi-turn conversations} 
To emulate realistic back-and-forth dialogue, we sample general-topic conversations from LMSYS-Chat-1M~\cite{zheng2023lmsys} following previous studies~\cite{zhao2025llms,li2023long,bai2024mt} and strip any concept-related lines to avoid conflicts (Appendix \S\ref{sup:mt-turn}).
\paragraph{Images for personalized query} For personalized query images for recognition, we follow the image collection guidelines, ensuring that injection and evaluation images are strictly disjoint.
For preference-grounded VQA, human annotators collect images from approved sources associated with 30 predefined subdomains, ensuring relevance and diversity within each domain (Appendix \S\ref{sup:query_guideline}).

\subsection{Query Design}
\label{subsec:question_design}


\paragraph{Overall process}

The query design process consists of two main stages (Figure~\ref{fig:construction}):
(1) a generation stage, where a model drafts candidate queries and human annotators refine them to produce initial query sets; and
(2) a quality evaluation stage, where models first flag suspected low-quality queries, and human annotators validate the flags and review all queries.
We use Ovis2-34B~\cite{lu2024ovis} in the generation stage, and Ovis2-34B, InternVL2.5-78B~\cite{chen2024internvl}, and Qwen2.5-VL-72B~\cite{bai2025qwen2.5} in the evaluation stage. 
Although \mmpb{} is designed as a multiple-choice VQA, all queries are easily convertible into open-ended formats (Appendix Table~\ref{tab:generated_samples}). As in \S\ref{subsec:data_collection}, all concept names are replaced with \sks{}.


\paragraph{Categories} For query categorization, we follow the \textit{Personalization Criteria} (\S\ref{sec:sec.2}) and define three types of queries:  
(1) \textit{Awareness},  
(2) \textit{Appropriateness}, and  
(3) \textit{Coherency}.  
Each query type is applied to recognition tasks across all concept categories, and to preference-grounded tasks for human concepts, resulting in 15 evaluation tasks.
The \textit{Awareness} type tests whether the model can detect the presence of a personalized concept in an image. Positive images are used, where the correct answer to queries like “Is \sks{} in the image?” should be “yes.” We further distinguish between single-entity and multi-entity cases, based on whether other entities co-appear with the concept.
The \textit{Appropriateness} type evaluates whether the model can correctly suppress references to a personalized concept when it is not contextually appropriate. Negative images are used, and the correct answer should be “no.” For animal concepts, negative samples are further categorized into same-species and different-species examples to analyze the impact of hard negatives. This is feasible because species boundaries are visually and semantically well-defined.
The \textit{Coherency} type assesses whether the model can produce coherent and context-appropriate responses about the concept (\eg “What clothes is \sks{} wearing?”).
We also evaluate \textit{Persistency} by introducing multi-turn conversations.

\paragraph{Quality control}

\begin{wrapfigure}{r}{0.3\textwidth} 
  \vspace{-1.25em}
  \hspace{-0.5em}
  \centering
  \includegraphics[width=0.3\textwidth]{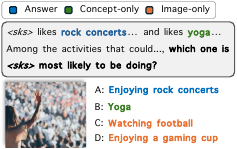}
    \caption{Example of quality control for a \textit{Coherency}-type query with concept-only and image-only distractors.}
  \label{fig:wrapped}
  \vspace{-0.8em}
\end{wrapfigure}

We aim to filter out options that trigger positional or affirmative biases or are solvable by text alone~\cite{liu2024mmbench,robinson2022leveraging,zhao2024difficult}. Throughout query generation and evaluation, we follow task-specific guidelines to ensure high quality\footnote{
Each query is reviewed by at least three human annotators, and only queries with majority agreement (\ie at least 2 out of 3 annotators) are accepted. Please see the guidelines in Appendix~\S\ref{sup:guide_query_gen_human} and~\S\ref{supsec:human_guide_eval}. 
}.
For the \textit{Awareness} and \textit{Appropriateness} types, we instruct both the model and human annotators to generate queries targeting concept presence in each image. To mitigate potential yes/no bias~\cite{zhao2024difficult}, we include both affirmative and negative formulations (\eg “Is \sks{} not present in this photo?”).
In the \textit{Coherency} type, we adopt a 4-option MCQ format, addressing two major confounding factors: (i) \textbf{concept-only solvability} and (ii) \textbf{image-only solvability}. 
As shown in Figure~\ref{fig:wrapped}, at least one distractor (\eg "Yoga") is aligned with the \sks{} but not with the image, preventing concept-only solvability. Other distractors (\eg "Watching football") are visually plausible but incorrect without considering the concept, addressing image-only solvability. This design encourages joint reasoning over both the image and the concept, with choices shuffled to eliminate positional bias~\cite{robinson2022leveraging}. Blind tests with text-only experiments are provided in Appendix Table~\ref{suptab:blind}, further supporting our rigorous query design.


\section{Benchmarking Results}
\label{sec:benchmark}

\subsection{Experimental Setup}

\paragraph{Configuration} To simulate real-world personalization challenges, we design our experiments centered on two key constraints: \textit{cold-start} and \textit{multi-turn} adaptation. For the cold-start problem~\cite{zhao2025llms,yu2021personalized} (\ie initiating personalization with minimal prior information), we utilize \textit{moderate} description for text-based injection, and two reference images for image-based injection with a single concept injection.
We also evaluate both in 0-turn (\ie without multi-turn conversation) and 10-turn conversation settings. We ensure that the 10-turn conversation is applied consistently across all models. We follow prior multiple-choice VQA tasks~\cite{romero2024cvqa,antol2015vqa,zhu2016visual7w} and use overall accuracy as our main evaluation metric.

\begin{figure}
    \centering
    \includegraphics[width=1\linewidth]{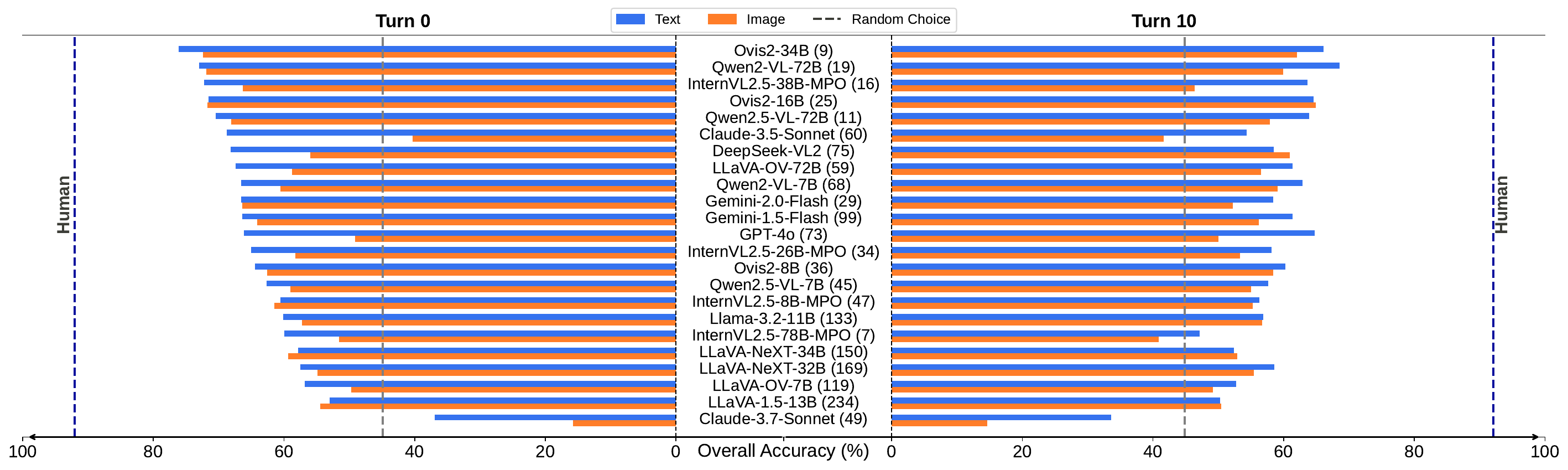}
    \caption{
    Evaluation results of 23 VLMs on \mmpb{} under 0-turn and 10-turn settings. Model names are followed by their average ranks across eight general-purpose multi-modal benchmarks\protect\footnotemark.} 
    \label{fig:main_results}
    \vspace{-1em}
\end{figure}

\footnotetext{\href{https://huggingface.co/spaces/opencompass/open_vlm_leaderboard}{Hugging Face Open VLM Leaderboard}, accessed on May 15, 2025.}

\paragraph{Models} 

We select models to cover major VLM families following previous benchmarks~\cite{gavrikov2025can, chen2025megabench}, based on public availability, and widespread usage. We evaluate 23 models across open- and closed-source families: open-source includes InternVL2.5 (8–78B)~\cite{chen2024internvl,chen2024far,chen2024expanding}, Ovis2 (8–34B)\footnote{To ensure the model does not benefit from self-generated content, we confirm that Ovis2-34B underperforms on its own generated descriptions compared to human-written \textit{Moderate} descriptions (Appendix \S\ref{supsubsec:ovis2}).}~\cite{lu2024ovis}, LLaVA (13–72B)~\cite{liu2023visual,liu2024improved,li2024llava}, Qwen-VL (7–72B)~\cite{wang2024qwen2,bai2025qwen2.5}, DeepSeek-VL2~\cite{wu2024deepseek2}, and Llama-3.2-11B~\cite{grattafiori2024llama3}; closed-source includes Claude-3.5/3.7-Sonnet~\cite{bai2022constitutional}, Gemini-1.5/2.0-Flash~\cite{team2024gemini1.5}, and GPT-4o~\cite{achiam2023gpt}. 
\subsection{Overall Results}
\label{subsec:main_results}

Figure~\ref{fig:main_results} shows overall personalization performance on \mmpb{} across 23 widely used VLMs, with the human evaluator achieving an average of 92.1\%, thereby establishing the upper bound (Appendix \S\ref{sup:humaneval}). Across all experiments, text-based concept injection consistently outperforms image-based injection, achieving average accuracies of 63.8\% vs. 57.8\% in the 0-turn, and 57.9\% vs. 52.6\% in the 10-turn setting. The difficulty becomes more pronounced under extended interactions. Most models experience substantial performance drop when moving from the 0-turn to the 10-turn setting, indicating challenges in maintaining personalized responses over time. Notably, averaged across all experimental settings, closed models underperform open ones (51.4\% vs. 59.9\%) despite strong results on general VQA benchmarks~\cite{chen2024mega,yue2024mmmu}. Table~\ref{tab:detail_result} presents task- and concept-level performance comparisons between model types. Overall, both model types exhibit pronounced multi-turn degradation.  

To further examine whether \mmpb{} genuinely measures personalization rather than merely reflecting general VLM strength, we conducted additional experiments on LLaVA-1.5-13B with two personalization strategies: (i) soft prompt tuning following the Yo’llava~\cite{nguyen2024yo} approach, and (ii) personalized LoRA fine-tuning~\cite{zhang2024personalized}. As detailed in Appendix~\S\ref{sup:pfinetuning}, both methods lead to clear improvements over the baseline, with LoRA providing the strongest gains. Overall, these results validate that MMPB is sensitive to personalization techniques and effectively captures personalization performance beyond general VLM ability.

\begin{table}[t]
  \centering
    \caption{Task-wise performance of open vs. closed models, averaged over image/text-based injections.}%
  \setlength{\tabcolsep}{14pt}
  \resizebox{0.95\textwidth}{!}{ 
  \renewcommand{\arraystretch}{0.8}

    \begin{tabular}{lcccccc}
      \toprule
      & & \multicolumn{1}{c}{\textbf{Preference}} & \multicolumn{4}{c}{\textbf{Recognition}} \\
      \cmidrule(lr){3-3} \cmidrule(lr){4-7}
      \textbf{Model} & \textbf{Turn} & Human & Human & Animal & Object & Character \\
      \midrule
      Open   & 0  & 52.8 {\scriptsize$\pm$ 5.9}  & 74.3 {\scriptsize$\pm$ 11.1} & 69.2 {\scriptsize$\pm$ 13.7} & 75.7 {\scriptsize$\pm$ 18.1} & 67.3 {\scriptsize$\pm$ 11.5} \\
      Closed & 0  & 37.0 {\scriptsize$\pm$ 18.0} & 68.9 {\scriptsize$\pm$ 21.0} & 73.4 {\scriptsize$\pm$ 9.0}  & 79.1 {\scriptsize$\pm$ 16.0} & 65.3 {\scriptsize$\pm$ 17.4} \\
      \midrule
      Open   & 10 & 46.0 {\scriptsize$\pm$ 5.1}  & 69.0 {\scriptsize$\pm$ 12.4} & 69.8 {\scriptsize$\pm$ 8.3}  & 72.0 {\scriptsize$\pm$ 14.0} & 61.6 {\scriptsize$\pm$ 10.9} \\
      Closed & 10 & 34.7 {\scriptsize$\pm$ 15.7} & 61.7 {\scriptsize$\pm$ 16.0} & 66.2 {\scriptsize$\pm$ 10.6} & 67.3 {\scriptsize$\pm$ 19.0} & 56.9 {\scriptsize$\pm$ 15.9} \\
      \bottomrule
    \end{tabular}}
    \vspace{-0.5em}
    \label{tab:detail_result}
\end{table}


\section{Challenges in VLM Personalization}
\label{subsec:challenges}

The results presented in \S\ref{subsec:main_results} demonstrate that, overall, recent VLMs remain highly vulnerable to personalization tasks. To systematically investigate which challenges hinder effective handling of user-specific queries, we organize our analysis around the following key questions:

\begin{itemize}
    \item Which specific tasks pose barriers to effective VLM personalization? (\S\ref{subsubsec:preference_grounded_vqa} and \S\ref{subsubsec:systematic_error_patterns})
    \item What makes closed-source models struggle with personalization? (\S\ref{subsubsec:resistance_closed_models})
    \item  How do injection modality and granularity impact VLMs? 
 (\S\ref{subsubsec:suboptimal_visual_concept_injection})
    \item Can VLMs follow a specific concept in multi-concept or long-context settings? (\S\ref{subsubsec:multi_turn_degradation})
\end{itemize}

\subsection{Preference-Grounded Reasoning}
\label{subsubsec:preference_grounded_vqa}

\begin{figure*}[!t]
  \centering
  \begin{minipage}[t]{0.48\textwidth}
    \centering
    \includegraphics[width=0.825\linewidth]{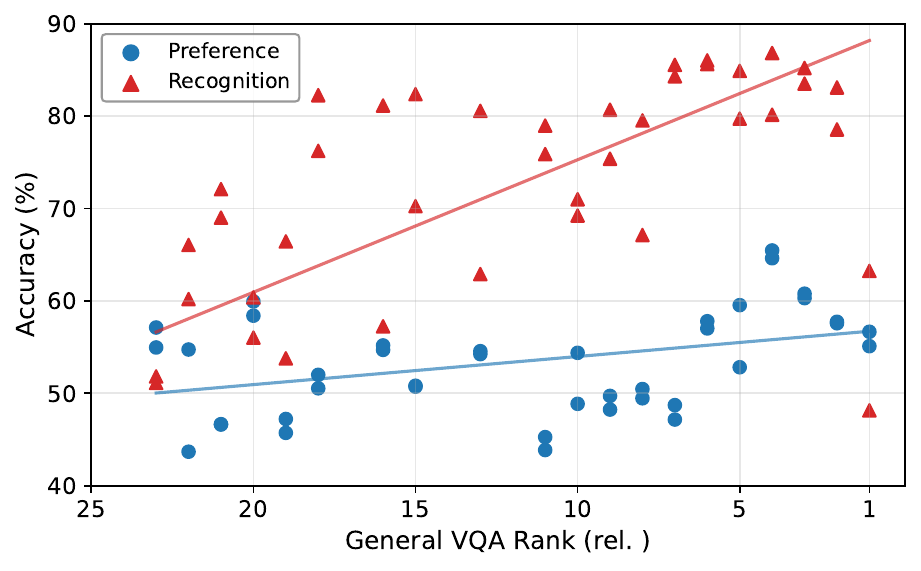}
    \caption{Performance gap between preference-grounded and recognition VQA tasks in VLMs.}
    \label{fig:chel_pref}
  \end{minipage}\hfill
  \begin{minipage}[t]{0.48\textwidth}
    \centering
    \includegraphics[width=0.815\linewidth]{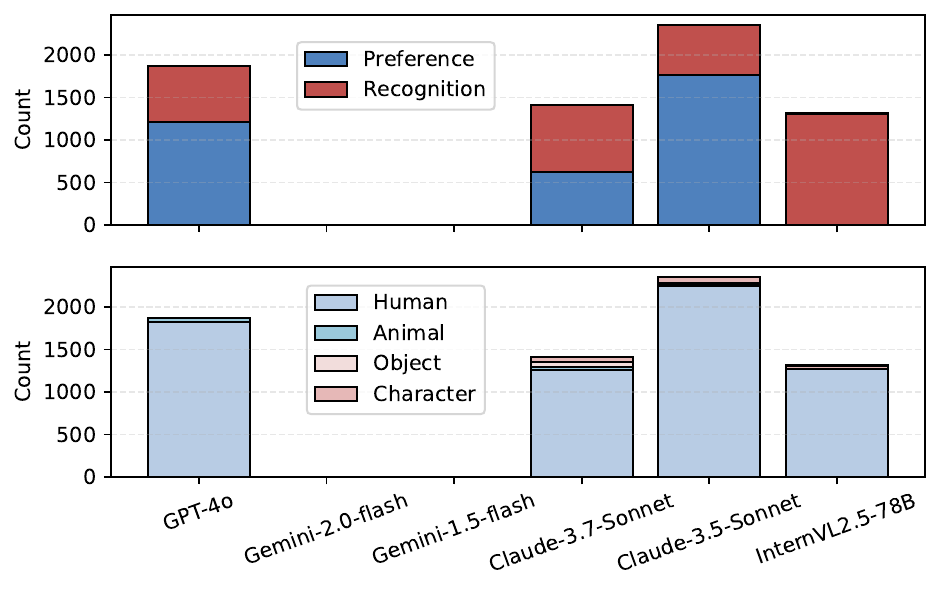}
    \caption{Refusal counts of models across task and concepts under image-based injection.}
    \label{fig:chel_closed}
  \end{minipage}
\end{figure*}

Personalized VLMs should be able to infer user preferences from images and incorporate them into its responses. We split the evaluation into recognition and preference tasks and order their accuracies by average relative rank within general VQA, which comprises eight multi-modal reasoning benchmarks, including MathVista~\cite{lu2023mathvista}, MMMU~\cite{yue2024mmmu}, and MMBench~\cite{liu2024mmbench}. As shown in Figure~\ref{fig:chel_pref}, models that perform well on general tasks consistently struggle with preference-grounded queries. This disparity highlights a key limitation of current VLMs: their inability to generalize from general-purpose reasoning to user-centric inference. In contrast, recognition tasks show a stronger correlation with rankings, underscoring the unique difficulty posed by preference-based personalization.


\paragraph{Difference between recognition and preference-grounded VQA}
VLMs typically solve VQA tasks through a three-stage process: image perception, knowledge grounding (either pretrained or retrieved via RAG~\cite{lewis2020retrieval}), and answer generation. In \mmpb{}, recognition tasks follow this pipeline with minimal uncertainty, using deductive reasoning on injected concepts via in-context learning. For example, given “Is \sks{} in this picture?”, the model matches visual features and answers “Yes.” By contrast, preference-grounded VQA demands abductive reasoning, integrating user intentions and tastes with scene understanding, \eg in “Which activity here do you think \sks{} would enjoy most?” Current VLMs primarily focus on deductive tasks~\cite{liu2023visual}, which may be insufficient for ensuring user satisfaction in real-world scenarios. Future personalization frameworks may consider training VLMs to perform abductive reasoning, \eg via instruction tuning on preference-based tasks.

\begin{table}
\vspace{-1em}
  \centering
  \caption{Examples of evasive responses in human category tasks.}
  \small
  \setlength{\tabcolsep}{12pt}
  \renewcommand{\arraystretch}{0.25}
  \begin{tabular}{lc}
    \toprule
    \textbf{Model}  & \textbf{Example} \\
    \midrule
    \includegraphics[height=1em]{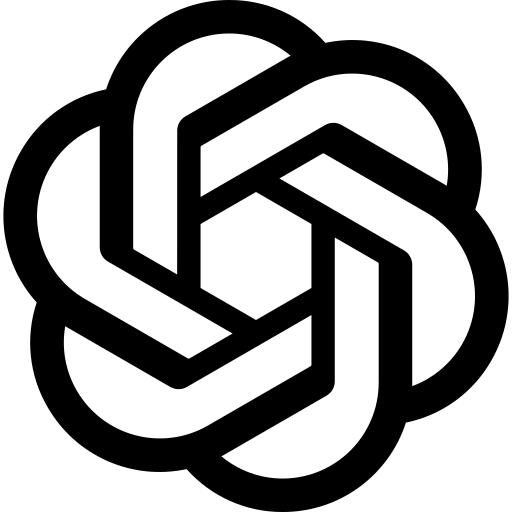} GPT-4o & Unknown  \\
    \includegraphics[height=1em]{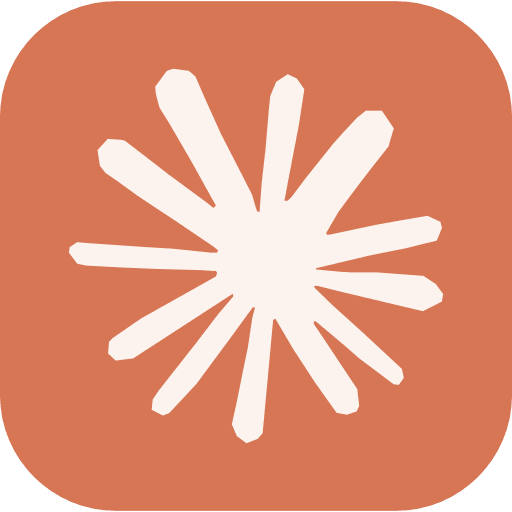} Claude-3.7-Sonnet&   Based on the image, I can only describe what's visible...  \\
    \includegraphics[height=1em]{figures/models/claude.png} Claude-3.5-Sonnet&   I cannot and should not identify or name specific features of the person...   \\
    \includegraphics[height=1em]{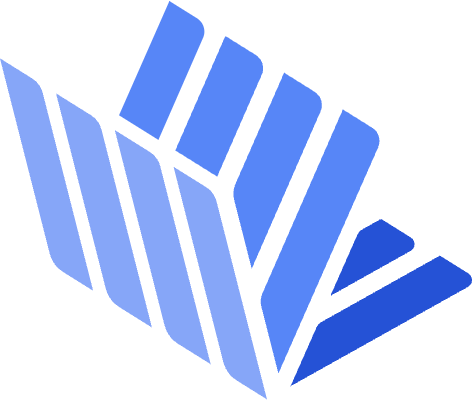} InternVL2.5-78B  &  I can't identify or recognize people in images. \\
    \bottomrule
  \end{tabular}
  \label{tab:model_examples}
\end{table}



\subsection{Resistance to Personalization}
\label{subsubsec:resistance_closed_models}

As shown in Table~\ref{tab:detail_result}, closed-source models consistently underperform on human-related queries. Notably, InternVL2.5-78B, which ranked first among 23 models on a general VQA benchmark, also exhibits weak personalization performance, both in the text- and image- based setting (Figure~\ref{fig:main_results}). We find that this is the only model in the InternVL family that is publicly accessible via a web-based chatbot, which may introduce additional safety filters that affect its behavior. To investigate their failures in detail, we examine their responses and find a consistent pattern of evasive behaviors.

To quantify this behavior, we count answers matching well-known refusal patterns in LLMs~\cite{refuse1,refuse2,refuse3,refuse4,refuse5}, such as \textit{``I'm sorry,'' ``I shouldn’t,’’} and \textit{``Unknown.’’}. As shown in Figure~\ref{fig:chel_closed}, all closed-source models, except Gemini, exhibit evasive responses in recognition and preference-grounded tasks. InternVL2.5-78B does so only in recognition. Notably, most evasive cases fall under the human category: out of 7,501 queries, up to 2,237 trigger evasive answers, suggesting that this behavior significantly contributes to suboptimal performance. Table~\ref{tab:model_examples} presents representative free-form responses, aligning with refusal patterns observed when LLMs are prompted with harmful queries~\cite{pasch2025llm}. Interestingly, with the upgrade from Claude-3.5-Sonnet to 3.7, the model increasingly produces descriptive outputs rather than issuing explicit refusals. Since these responses do not clearly match standard refusal patterns, we exclude them from the count. Further analysis (Appendix Box~\ref{response_explanation1} and \ref{response_explanation2}) shows that closed-source models treat identity-related queries as privacy-sensitive, resulting in conservative behavior aimed at avoiding unsafe outputs.

\paragraph{Personalization vs. Safety: Can they coexist?} This tradeoff between personalization and safety raises important questions for the AI community. Human perception is a critical component of user-level personalization, particularly in tasks that involve grounding visual inputs to individual users. While safety constraints are essential to prevent misuse, an overly cautious stance can hinder meaningful personalization, where distinguishing between users is a core requirement. This tension calls for deeper discussion around the boundary between safety and utility, and highlights the need for methods that enable secure yet effective handling of identity information in VLMs.

\subsection{Personalization Bias and Fine-Grained Discrimination}
\label{subsubsec:systematic_error_patterns}

\begin{figure}[t]
  \centering
    \includegraphics[width=.95\linewidth]{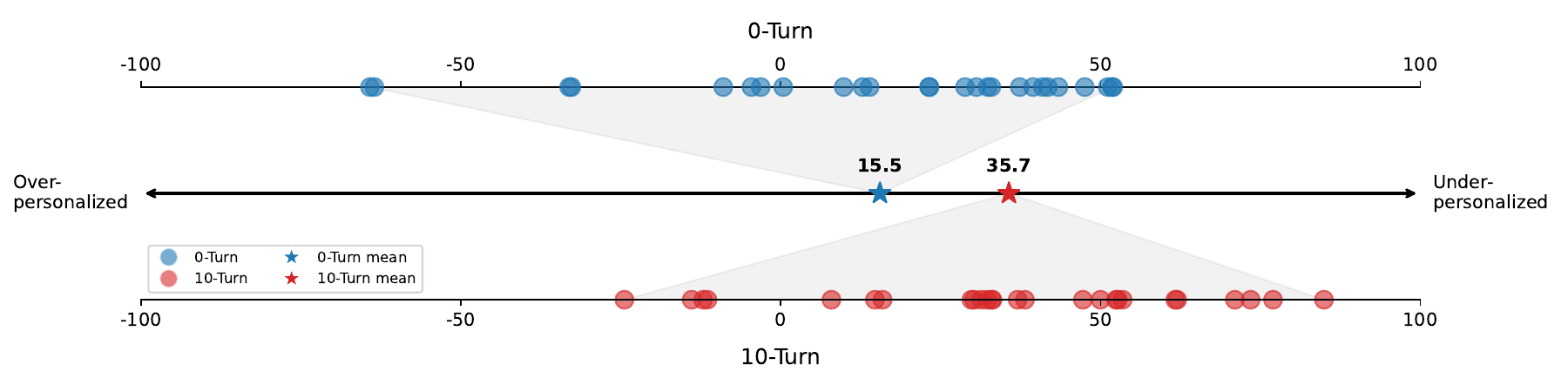} 
    \caption{Personalization bias across models, measured as \textit{Appropriateness}–\textit{Awareness}. Positive values indicate under-personalization (missed valid personalization), and negative values indicate over-personalization (affirming inappropriate inputs).}
    
    \label{fig:error_type}
\end{figure}

\paragraph{Under-personalized bias} 

Figure~\ref{fig:error_type} visualizes the personalization bias of models, measured as the difference between \textit{Appropriateness} and \textit{Awareness}. In both the 0-turn and 10-turn settings, models consistently lean toward rejecting personalization, performing better on \textit{Appropriateness} than on \textit{Awareness}, with 72 out of 92 cases falling in the under-personalization region. This indicates a systematic tendency to reject personalized concepts rather than affirm them. The bias is more pronounced in the 10-turn setting, suggesting that extended dialogue reinforces this behavior.

\begin{figure}
  \centering
  \begin{subfigure}[b]{0.495\textwidth}
    \centering
    \includegraphics[width=\linewidth]{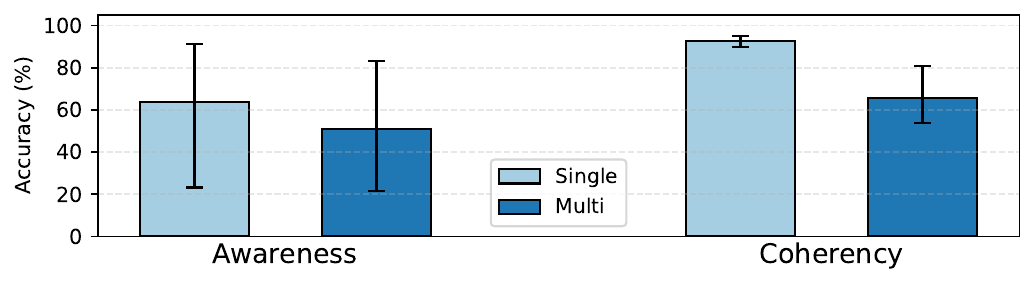}
    \caption{}
    \label{fig:sub2}
  \end{subfigure}
  \hfill
  \begin{subfigure}[b]{0.495\textwidth}
    \centering
    \includegraphics[width=\linewidth]{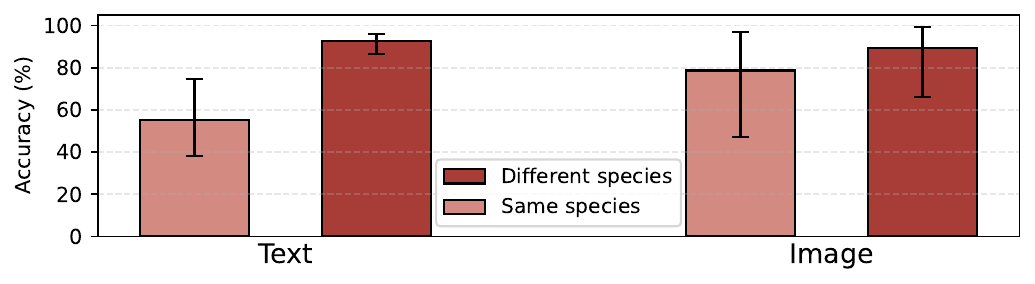}
    \caption{}
    \label{fig:sub3}
  \end{subfigure}
    \caption{Effects of image complexity: (a) Multiple entities on \textit{Awareness} and \textit{Coherency} tasks. (b) Hard-negative samples on \textit{Appropriateness} task. The error bars denote the first and third quartiles.}
  \label{fig:type_analysis}
  \vspace{-1em}
\end{figure}

\paragraph{Impact of multiple entities and hard-negatives} Figure~\ref{fig:sub2} compares performance on single- and multi-entity query images. The models exhibit comparable \textit{Awareness} scores, suggesting that concept detection remains stable. However, \textit{Coherency} drops substantially in response to multi-entity inputs, indicating difficulty in maintaining reasoning consistency as input complexity increases. To evaluate fine-grained \textit{Appropriateness}, we assess whether the models can distinguish target concepts from visually similar distractors in the animal category, using same-species (\eg Beagle–Beagle) and different-species (\eg Beagle–Shiba) instances as negative samples. As shown in Figure~\ref{fig:sub3}, text-based injection results in a substantial performance drop for hard negative samples, highlighting the challenges in making fine-grained distinctions. In contrast, image-based injection delivers robust performance, suggesting that visual cues offer stronger support for fine-grained discrimination.

\subsection{Visual Personalization}
\label{subsubsec:suboptimal_visual_concept_injection}

We conduct an in-depth analysis of how injection modality and content granularity affect model performance by comparing four levels of text granularity and three levels of image-based injection. We evaluate six representative models, including Qwen2/2.5-VL-7B, Ovis2-8/16B, LLaVA-OV-7B, and InternVL2.5-8B-MPO, all of which support multi-image input. Results are presented in Table~\ref{tab:modality_comparison}.
\paragraph{A few words are worth a thousand pixels}  
We highlight distinct trends in recognition performance across modalities. Surprisingly, \textit{Simple} textual injections, consisting of only three keywords, achieve comparable accuracy to the \textit{1-image} condition in 10-turn, suggesting that the use of a single image doesn’t provide much more benefit than minimal text alone. Moreover, even with five reference images, the performance does not exceed that of \textit{Moderate} text injection, highlighting the limited utility of visual examples in recent VLMs. Given the central role of visual personalization in future VLMs, further research should explore strategies that effectively harness these visual cues for personalization.

\begin{table}[t]
  \caption{Recognition accuracies under different granularity levels of concept injection.}
  \centering
  \setlength{\tabcolsep}{10pt}
  \renewcommand{\arraystretch}{0.75}
  \resizebox{0.95\textwidth}{!}{
    \begin{tabular}{lccccccc}
      \toprule
      & \multicolumn{4}{c}{\textbf{Text-based Injection}} & \multicolumn{3}{c}{\textbf{Image-based Injection}} \\
      \cmidrule(lr){2-5} \cmidrule(lr){6-8}
      \textbf{Turn} & Simple & Moderate & Detailed & Extended & 1 image & 2 images & 5 images \\
      \midrule
      0  & 68.8 {\scriptsize$\pm$ 5.2}  & 77.1 {\scriptsize$\pm$ 6.7}  & \textbf{79.3} {\scriptsize$\pm$ 6.7} & 78.3 {\scriptsize$\pm$ 6.8} & 71.5 {\scriptsize$\pm$ 8.3} & 72.2 {\scriptsize$\pm$ 9.9} & \textbf{73.3} {\scriptsize$\pm$ 10.5} \\
      10 & 67.4 {\scriptsize$\pm$ 4.7}  & 72.0 {\scriptsize$\pm$ 7.2}  & \textbf{73.5} {\scriptsize$\pm$ 7.9} & 72.1 {\scriptsize$\pm$ 8.2} & 67.9 {\scriptsize$\pm$ 6.2} & 69.2 {\scriptsize$\pm$ 8.4} & \textbf{70.1} {\scriptsize$\pm$ 9.4} \\
      \bottomrule
    \end{tabular}
  }
  \label{tab:modality_comparison}
\end{table}

\begin{figure}
  \centering
  \begin{subfigure}[b]{0.329\textwidth}
    \centering
    \includegraphics[width=0.9\linewidth]{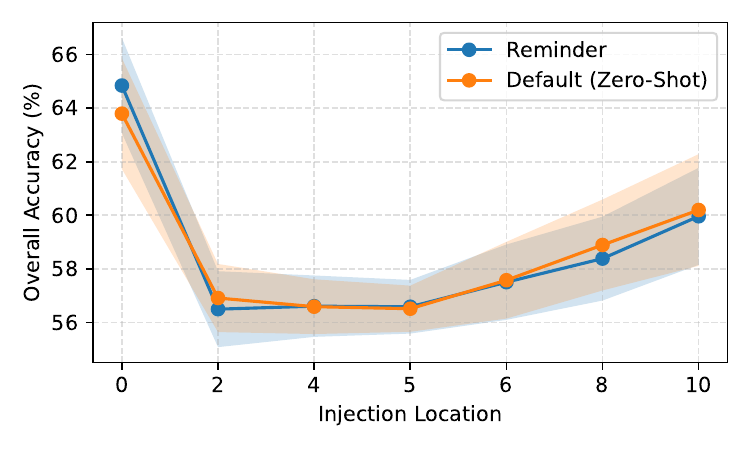}
    \caption{}
    \label{fig:lim}
  \end{subfigure}
  \hfill
  \begin{subfigure}[b]{0.329\textwidth}
    \centering
    \includegraphics[width=0.9\linewidth]{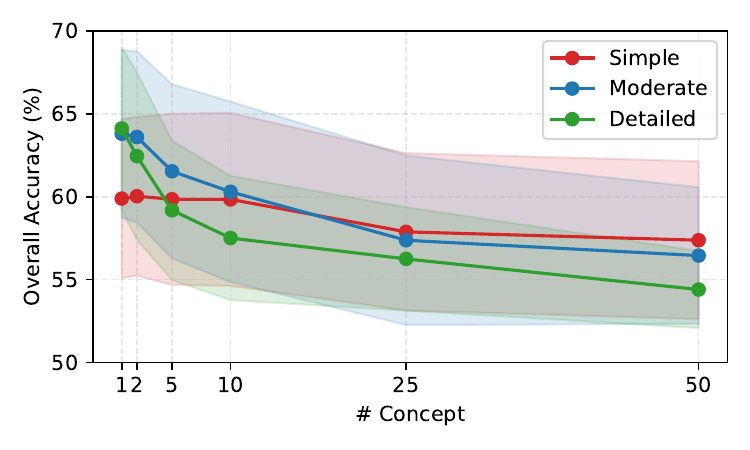}
    \caption{}
    \label{fig:multi_concept}
  \end{subfigure}
    \hfill
  \begin{subfigure}[b]{0.329\textwidth}
    \centering
    \includegraphics[width=0.9\linewidth]{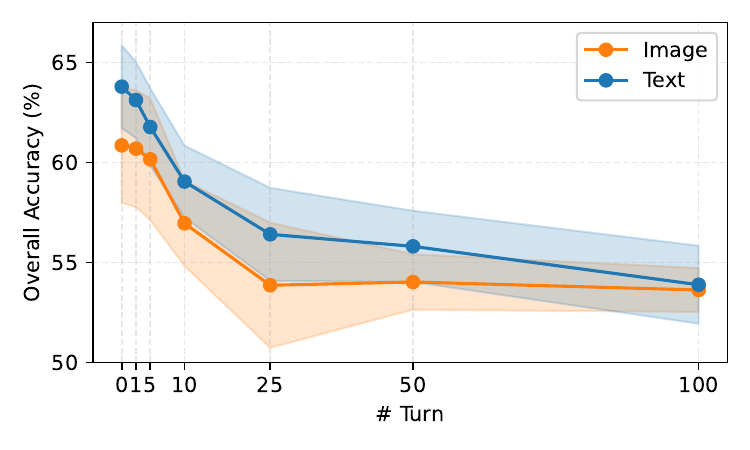}
    \caption{}
    \label{fig:multi_turn_overall}
  \end{subfigure}

    \caption{VLMs under long-context scenarios. (a) Concept most lost midway conversation. (b) Multiple concept injections on text-based injection. (c) Extended multi-turn dialogue up to 100 turns.}
  \label{fig:multi_turn}
  \vspace{-1em}
\end{figure}

\subsection{Lost Concepts in a Haystack}
\label{subsubsec:multi_turn_degradation}

Prior work shows that VLMs struggle with the needle-in-a-haystack problem, failing to retrieve key information from long contexts~\cite{wang2024multimodal,chen2024longvila,hengle2024multilingual}. Using six models in \S\ref{subsubsec:suboptimal_visual_concept_injection}, we evaluate concept retrieval in diverse long-context settings. Across all settings, personalization failures stem less from input design, and more from fundamental limits in long-range memory and relevance filtering.

We begin with 10-turn dialogues in which concepts are injected at different positions (Figure~\ref{fig:lim}). We also apply a \textit{Reminder} strategy~\cite{zhao2025llms,wu2025effectively}, which cues concept recall in the final turn (Appendix~\S\ref{supsec:prompting}). As a result, regardless of the prompting methods, models often forget concepts located near the midpoint, reflecting the “lost-in-the-middle” effect~\cite{wu2024visual,liu2023lost,hsieh2024found,zhang2024found} and indicating a positional bias in attention. We also evaluate multi-concept inputs with up to 50 entities and varying description granularity in the 0-turn setting, placing the target concept near the middle (\eg 6th of 10). In Figure~\ref{fig:multi_concept}, the accuracy consistently decreases, with sharper drops observed in more detailed descriptions. These findings similarly indicate that VLMs struggle to isolate relevant information embedded mid-context, particularly when multiple entities are present. In extended dialogues of up to 100 turns (Figure~\ref{fig:multi_turn_overall}), performance for both text- and image-based inputs deteriorates sharply after 5 turns, reaffirming that VLMs’ context-tracking ability degrades significantly with increased input length~\cite{zhao2025llms}.


\section{Related Works}
\paragraph{Large Vision-Language Models} 
Large language models (LLMs)~\cite{nousyi34b,phi2,jiang2023mistral,bi2024deepseek,young2024yi,jiang2024mixtral,team2023internlm,chatgpt,touvron2023llama}, built on extensive world knowledge, have grown rapidly across diverse domains. 
Leveraging this progress, large Vision–Language Models (VLMs)~\cite{instructblip,li2023blip,zhang2022glipv2,fang2023eva,ye2023mplug,zhang2023internlm, zhu2025internvl3,team2024gemini1.5,lu2024ovis} have emerged to connect visual and linguistic modalities. 
By seamlessly bridging vision and language, VLMs have revolutionized AI, 
driving breakthroughs in tasks such as visual question answering~\cite{gurari2018vizwiz,marino2019ok,schwenk2022okvqa,wu2023q,yue2024mmmu}, 
image captioning~\cite{krishna2017visual,kazemzadeh2014referitgame,li2023seed,urbanek2023picture}, 
and optical character recognition (OCR)~\cite{mishra2019ocr,lu2023mathvista}. 
Despite their potential as interfaces connecting AI systems with individual users in real-world scenarios, current benchmarks lack the means to evaluate the personalization capabilities of VLMs. 
To address this gap, we propose \mmpb{}, the first benchmark designed for VLM personalization, establishing a foundation for future research in personalized vision–language modeling. 
Unlike conventional VQA or captioning datasets, our benchmark highlights the user-dependent dimension of multi-modal reasoning, which has been largely underexplored in prior work.

\textbf{Personalization in AI systems.} 
As AI systems become more widely adopted across various domains such as AI assistants (e.g., psychological counseling~\cite{zhong2024memorybank}, smartphone~\cite{wang2024crafting}, housekeeping~\cite{han2024llm}, medical assistance~\cite{zhang2023memory}, and GUIs~\cite{hong2024cogagent}), personalization has become crucial for enhancing usability and user satisfaction. 
In the context of large language models (LLMs), various techniques such as prompting~\cite{salemi-etal-2024-lamp, salemi2024optimizationmethodspersonalizinglarge, zhuang2024hydra}, fine-tuning~\cite{ning2024userllmefficientllmcontextualization, zhang2024personalized, zhong-etal-2021-useradapter}, and alignment~\cite{kirk2024prism, chen2025pad, jang2023personalizedsoupspersonalizedlarge} have been explored. 
In contrast, while large Vision-Language Models (VLMs)~\cite{achiam2023gpt, team2024gemini1.5} have demonstrated strong performance on multi-modal tasks~\cite{liu2024mmbench, chen2024mmstar, park2025second, huang2024opera}, their personalization capabilities remain underexplored. 
Existing approaches~\cite{nguyen2024yo,alaluf2024myvlm} primarily focus on prior adaptation to personalizable concepts while overlooking cold-start scenarios, thereby limiting their applicability in real-world settings. Distinct from prior work, \mmpb{} is the first to systematically evaluate personalization in VLMs through hierarchical concept injection, enabling a structured assessment of how models internalize and maintain user-specific concepts across varying levels of interaction complexity.

\section{Discussion}

In this paper, we introduce \mmpb{}, the first benchmark for large Vision-Language Model (VLM) personalization, covering diverse personalizable concepts and task types, including preference-grounded VQA. Our human-in-the-loop dataset construction highlights its potential for future extensions, while our query filtering strategy ensures data consistency and quality. Extensive analysis shows that recent VLMs remain suboptimal across several personalization dimensions: limited preference-grounded reasoning, a tendency toward refusal, and ineffective use of visual cues. Although we focus on static appearance and preferences, future work should explore dynamically evolving user traits. Also, while we focus on VQA concerning concept presence and preference following, future benchmarks should cover application‐driven tasks like personalized captioning and personalized robots. We hope \mmpb{} encourages further progress in VLM personalization, accelerating the development of human-centric, real-world applications.

\paragraph{Limitations}
\mmpb{} focuses on evaluating VLMs’ ability to recognize and reason about fine-grained, static appearances and preferences. In practice, however, personal appearance and inclinations are rarely fixed: people change hairstyles, update their fashion, and see their own tastes evolve over time. As the first extensive benchmark for VLM personalization, \mmpb{} establishes a solid foundation—but we encourage future work to introduce dynamic concept updates (\eg style changes, evolving preferences) so that models can be assessed on their ability to track and adapt to real-world, time-varying identities.

A second limitation concerns the scope of our evaluation framework. In this work, we defined three core properties—\textit{Awareness}, \textit{Appropriateness}, and \textit{Coherency}—to capture key failure modes of personalized VLMs (alongside \textit{Persistency} with multi-turn conversations). Consequently, \mmpb{}’s assessments are necessarily formalized around these properties, rather than downstream applications. Yet personalized VLMs hold great promise for tasks such as tailored image captioning or personalized actions in robots that bridge individuals and AI. We anticipate that future benchmarks will build on \mmpb{} by measuring model performance in these real-world personalization scenarios.

\bibliography{main}

\appendix
\newpage
\part{Supplementary Material}

\etocsetnexttocdepth{subsection}
\localtableofcontents

\newpage

\section{Broader Discussion}



\subsection{Rationale Behind the Four Concept Categories}
We selected \textit{Human} because end users and the people around them (\eg family, friends, colleagues) represent the most common and socially salient personalization targets. The \textit{Animal} category reflects the growing importance of pets and other animals in users’ lives, enabling tailored interactions such as care reminders or activity suggestions. We include \textit{Object} to capture the wide range of personal belongings that smart-home devices and personalized robots must recognize and manage on behalf of the user. Finally, \textit{Character} covers virtual personas and narrative figures in VR, and metaverse, ensuring that VLMs can personalize experiences for users’ custom avatars and digital contexts.


\subsection{Additional Related Works}
\label{sup:add_rw}
\paragraph{VLM Personalization} 
Although VLM personalization remains an underexplored yet important area, several seminal works have recently emerged. 
MyVLM~\cite{alaluf2024myvlm} augments pretrained models such as BLIP-2~\cite{li2023blip} and LLaVA~\cite{liu2023visual} with external concept heads, enabling recognition of user-specific concepts and supporting personalized captioning and VQA. 
Yo’LLaVA~\cite{nguyen2024yo} builds on LLaVA to learn personalized concepts from only a handful of images, embedding them into special tokens to facilitate user-specific queries and conversations. 
MC-LLaVA~\cite{an2024mc} extends beyond single-concept personalization by introducing the first framework for multi-concept personalization, allowing models to jointly learn and reason about multiple user-specific concepts. 
PLVM~\cite{pham2024personalized} employs an Aligner module to encode reference images online, enabling the incorporation of new personalized concepts without additional fine-tuning costs. 
Finally, RAP~\cite{hao2025rap} leverages an external database and retrieval mechanisms to enhance personalization in multimodal LLMs, allowing real-time concept editing and dynamic updates that broaden applicability in real-world scenarios. 
Together, these works demonstrate the promise of personalized VLMs but also leave open challenges such as dynamic updates and long-term persistency. In particular, methods like Yo’LLaVA and MC-LLaVA depend on large sets of negative images to avoid concept confusion, raising data and training costs in practice.

\subsection{Ethical Consideration}
\label{supsub:ethic}
Identifying specific individuals or making unwarranted claims about their personal preferences has been extensively addressed by various alignment techniques. In our experiments, we likewise observed persistent refusal responses from both GPT-4o and the Claude family of models. Consequently, for the human category, \mmpb{} makes use of the human images provided by MC-LLaVA~\cite{an2024mc}, replacing each person’s real name with \sks{} to prevent pretraining data leakage and to avoid identifying private individuals wherever possible.

\section{Licensing}
\mmpb{} is licensed under CC BY-NC-SA 4.0. If you wish to access and use our dataset, you must understand and agree that it is for research purposes only and may not be used for any commercial or other purposes. Users assume all responsibility for any consequences arising from unauthorized use or redistribution. 

We do not own the copyrights to any movie stills or other non–Creative Commons images. We grant researchers access to these images on the condition that they acknowledge and respect the rights of the original copyright holders. We fully respect and honor the copyrights of the original authors.

If any original author requests that an image be removed, please contact us at \texttt{jake630@snu.ac.kr} or open an issue. We will then adjust our distribution method to provide only the image URLs.

\section{Annotation Process}
\label{sup:annotation}
In this section, we introduce our guidelines for both models and human annotators, following the three-stage data-construction process described in \S\ref{sec:construct}.

\subsection{Data Collection}
\label{sup:data_collection}
\includepdf[
scale=0.9,pages=3,pagecommand=\subsubsection{Images for Concept Injection}\label{sup:guide_concept_injection}]{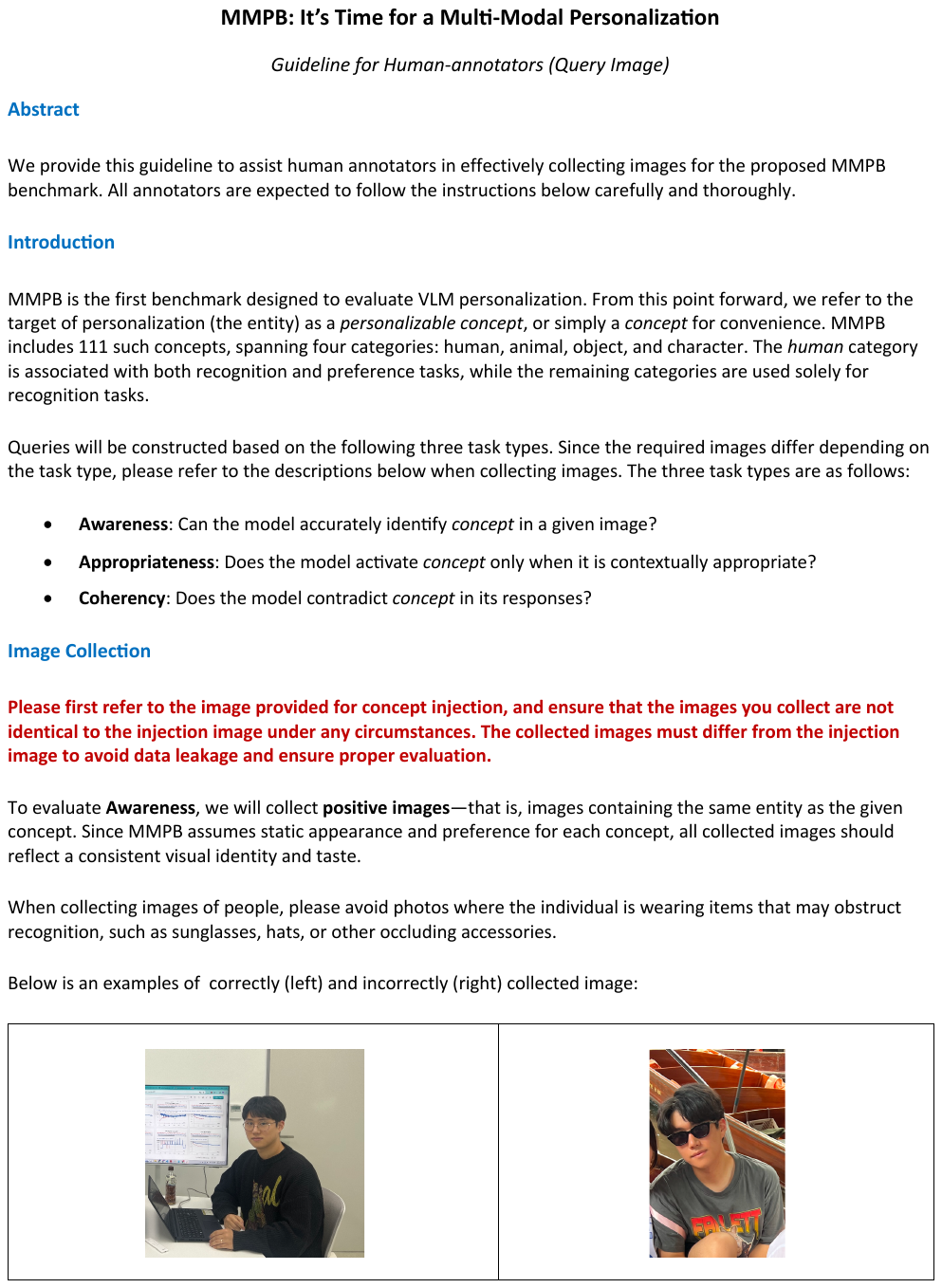}

\subsubsection{Text-based concept injection}
\label{sup:text_description}
Below is the prompt for generating four levels of textual descriptions for each concept (Human category for example).

\begin{tcolorbox}[colback=gray!10, colframe=black!80, boxrule=0.5pt, arc=2pt, title=Description generation guidelines for model (Simple)]

Analyze the five provided images, all depicting the same person. Based on your analysis, generate 3 concise keywords that best summarize their identity, appearance, or distinguishing features. \\

Provide your answer strictly in the format: <sks> is <keyword1>, <keyword2>, <keyword3>.
\end{tcolorbox}

\begin{tcolorbox}[colback=gray!10, colframe=black!80, boxrule=0.5pt, arc=2pt, title=Description generation guidelines for model (Moderate)]

Describe the person in the five images, highlighting their key physical traits and distinguishing features in a single sentence. \\

Name the person as <sks> and ensure the description uses '<sks>' naturally throughout.
\end{tcolorbox}

\begin{tcolorbox}[colback=gray!10, colframe=black!80, boxrule=0.5pt, arc=2pt, title=Description generation guidelines for model (Detailed)]

Carefully observe the person across all five images and provide a single, unified description of <sks>.  \\

Name the person as <sks> and ensure the description uses '<sks>' naturally throughout.  \\
Focus on consistent traits rather than describing each image separately. \\

Describe <sks>'s physical appearance, clothing style, and distinguishing features in a single paragraph.
\end{tcolorbox}

\begin{tcolorbox}[colback=gray!10, colframe=black!80, boxrule=0.5pt, arc=2pt, title=Description generation guidelines for model (Extented)]

Carefully examine all five images featuring the {attribute} referred to as <sks>. \\

Then, synthesize a comprehensive, unified description of <sks> that integrates consistent visual traits and personality cues across the full image set. 
Do **not** describe the images one by one—focus instead on what remains **visually and stylistically consistent** throughout. \\

Your output should consist of three rich, detailed paragraphs, each with a distinct focus: \\

(1) Physical Appearance: Describe <sks>’s body shape, height impression, facial structure, skin tone, hair style, and any recurring facial expressions or poses. 
Be specific and observational—mention anything from jawline shape to eyebrow thickness if consistent. \\

(2) Outfit and Accessories: Describe the style, color palette, textures, and function of <sks>’s clothing. 
Include details such as patterns, cuts, materials (e.g. denim, leather, silk), and whether the outfit feels practical, formal, whimsical, or character-defining. If <sks> wears accessories (glasses, jewelry, belts, bags), describe them and their potential significance. \\

(3) Distinguishing Visual Features or Vibe: Identify the most memorable and defining features of <sks>—things a viewer would immediately associate with them. This could include signature items, unique silhouette, color theming, visual motifs, or an aura they convey (e.g., mischievous, stoic, ethereal). 
Think of what makes <sks> visually unmistakable. Always use the name <sks> naturally and repeatedly in the description, and maintain a confident, analyst tone.
\end{tcolorbox}
\newpage

Please note that the model‐generated descriptions exhibited three issues: 1) omission of the \sks{} placeholder; 2) descriptions that were overly specific to individual images; and 3) failure to accurately convey the intended concept. To address these problems, human annotators revised each set of descriptions—reintroducing the \sks{} placeholder, generalizing the descriptions, and ensuring conceptual accuracy.

\subsubsection{Keywords for preferences} 

For human preference keywords, we first defined a hierarchy of subdomains under each of five main domains, assigning six distinct topic areas to every subdomain. To build a comprehensive keyword set for augmenting diverse preferences, we then prompted GPT-4o with the following instruction:

\begin{tcolorbox}[colback=gray!10, colframe=black!80, boxrule=0.5pt, arc=2pt, title=Prompt for generating preference keywords]

Using the domains and subdomains shown in the image above, generate 6–10 representative keywords for each subdomain. These keywords will later serve as “likes” or “dislikes” traits for a given person. For example:\\
- Entertainment → Concerts subdomain might yield keywords like “rock concerts,” “jazz festivals,” etc.\\
- Travel → Food subdomain might yield keywords like “fine-dining,” “street food tours,” “local cuisine tastings,” etc.\\
- Lifestyle → Food subdomain might yield keywords like “high-protein diet,” “organic meal prep,” etc.\\

Please produce 6–10 concise, descriptive keywords for each of the following subdomains:\\
1. Entertainment: Concerts, Comedy, Festival, Performance, TV Shows, Movies  \\
2. Travel: Nature, Culture, Transport, Food, Leisure, Extreme Activities  \\
3. Lifestyle: Food, Beverage, Exercise, Wellness, Habits, Home  \\
4. Shopping: Audio, Beauty, Hobby, Tech, Lifestyle Goods, Kitchen  \\
5. Fashion: Casual/Formal, Street/Trendy, Outerwear, Accessories, Innovative, Sustainable
\end{tcolorbox}

Using the model-generated keywords, we randomly assign them to our fixed templates (see \S\ref{sec:construct}). Crucially, this process often produce contradictory \textit{like} / \textit{dislike} pairs. To resolve these conflicts, we task human annotators with applying the following guidelines to reconcile any preference contradictions.

\begin{tcolorbox}[colback=gray!10, colframe=black!80, boxrule=0.5pt, arc=2pt, title=Guideline for resolving preference contradiction]
You are given a master keyword set and descriptive preference keywords (likes and dislikes), broken down by subdomain. \\

For each of the following subdomains, identify any contradictory preference keywords—cases where a like directly conflicts with a dislike—and resolve each conflict by replacing one of the conflicting keywords with a non-conflicting keyword from the corresponding master set: \\

1. Entertainment: Concerts, Comedy, Festival, Performance, TV Shows, Movies  \\
2. Travel: Nature, Culture, Transport, Food, Leisure, Extreme Activities  \\
3. Lifestyle: Food, Beverage, Exercise, Wellness, Habits, Home  \\
4. Shopping: Audio, Beauty, Hobby, Tech, Lifestyle Goods, Kitchen  \\
5. Fashion: Casual/Formal, Street/Trendy, Outerwear, Accessories, Innovative, Sustainable  \\

Examples of contradictory preferences:  \\
- Shopping: likes: gaming accessories; dislikes: tech gadgets  \\
- Lifestyle: likes: Yoga; dislikes: stretching  \\
\end{tcolorbox}

The final keyword set for each individual is saved in JSON format, as shown in Listing~\ref{lst:angelina_concept}.

\newpage

\begin{lstlisting}[
  caption={Example JSON file of human preference (with reference image path).},
  label={lst:angelina_concept}
]
{
    "concept": "Angelina",
    "images": [
        "./human/train/Angelina/0.png",
        "./human/train/Angelina/1.png",
        "./human/train/Angelina/2.png",
        "./human/train/Angelina/3.png",
        "./human/train/Angelina/4.png"
    ],
    "preferences": {
        "entertainment": {
            "likes": [
                "rock concerts",
                "indie films"
            ],
            "dislikes": [
                "theater performances",
                "film festivals"
            ]
        },
        "travel": {
            "likes": [
                "wildlife safaris",
                "hot air balloon rides"
            ],
            "dislikes": [
                "historical site tours",
                "food tours"
            ]
        },
        "lifestyle": {
            "likes": [
                "meditation",
                "yoga"
            ],
            "dislikes": [
                "plant-based diet",
                "home cooking"
            ]
        },
        "shopping": {
            "likes": [
                "Bluetooth speakers",
                "wireless charging stations"
            ],
            "dislikes": [
                "ceramic pottery",
                "kitchen appliances"
            ]
        },
        "fashion": {
            "likes": [
                "hand-knitted sweaters",
                "silk scarves"
            ],
            "dislikes": [
                "handcrafted leather bags",
                "woven hats"
            ]
        }
    }
}
\end{lstlisting}

\newpage

\subsubsection{Multi-turn conversation}
\label{sup:mt-turn}
For evaluating \textit{Persistency}, we extract ten-turn conversations on general topics from LMSYS-1M~\cite{zheng2023lmsys}. These same dialogues are presented to every VLM under test. The conversations we select are listed below.

\begin{tcolorbox}[colback=gray!10, colframe=black!80, boxrule=0.5pt, arc=2pt, title=Default 10-turn conversation transcript (Excerpt)]

\begin{description}[style=nextline, leftmargin=2cm, labelsep=0.5cm]

\item[\textbf{User:}] hello what is your name
\item[\textbf{Assistant:}] Hello! My name is OpenAI Assistant. How can I assist you today?

\item[\textbf{User:}] You are a historian of the internet, please tell me why nobody uses Usenet, or newsgroup anymore ?
\item[\textbf{Assistant:}] Usenet and newsgroups were very popular in the early days of the internet, but their popularity has declined significantly in recent years. There are several reasons for this: 1. Lack of user-friendliness... 4. Limited access... they have largely been replaced by other platforms that are easier to use and offer more features.

\item[\textbf{User:}] What happened to the Javascript framework called Meteor, or MeteorJS, and what new framework or technologies did replace it ?
\item[\textbf{Assistant:}] Meteor is still an actively maintained and popular JavaScript framework... However, its popularity has declined... alternatives include: 1. React... 4. Node.js...

\item[\textbf{User:}] You are a database expert, please tell me why and when should I use MongoDB instead of a traditional database like postgresql ?
\item[\textbf{Assistant:}] MongoDB and PostgreSQL are both powerful database systems... scenarios where you might consider using MongoDB: 1. Handling large amounts of unstructured data... 4. Real-time analytics... Ultimately, the choice of database system will depend on your specific needs...

\item[\textbf{User:}] On a different topic now, why is the temperature colder when going up in the mountains ?
\item[\textbf{Assistant:}] The temperature drops as you go higher... due to a phenomenon known as adiabatic cooling... other factors... include thinner air and the presence of snow and ice.

\item[\textbf{User:}] Including their national waters, what are the five largest countries on earth ?
\item[\textbf{Assistant:}] The five largest countries on Earth by total area, including their national waters, are: 1. Russia... 5. Brazil...

\item[\textbf{User:}] Please do the same ranking, but this time incuding the exlcusive economic zones
\item[\textbf{Assistant:}] If we include the exclusive economic zones (EEZs)... 1. Russia — 25,064,100 km²... 5. Brazil — 8,515,767 km²...

\item[\textbf{User:}] Why are you called NAME\_1, a French name ?
\item[\textbf{Assistant:}] I apologize for any confusion, but I am not called NAME\_1 and I am not a person...

\item[\textbf{User:}] Where is based the company AI company Anthropic, and who are its founders ?
\item[\textbf{Assistant:}] Anthropic is an AI company that is based in San Francisco, California, USA. The company was founded in 2019 by NAME\_2 and NAME\_3... NAME\_4 is a prominent AI researcher... NAME\_5 is a software engineer...

\item[\textbf{User:}] Who are the investors of Anthropic ?
\item[\textbf{Assistant:}] Anthropic has raised funding from several notable investors, including: 1. NAME\_6... 5. NAME\_10... The company has also received funding from venture capital firms...

\end{description}

\end{tcolorbox}



### C.1.5 Query Images

## MMPB: It's Time for a Multi-Modal Personalization

*Guideline for Human-annotators (Query Image)*

### Abstract

We provide this guideline to assist human annotators in effectively collecting images for the proposed MMPB benchmark. All annotators are expected to follow the instructions below carefully and thoroughly.

### Introduction

MMPB is the first benchmark designed to evaluate VLM personalization. From this point forward, we refer to the target of personalization (the entity) as a *personalizable concept*, or simply a *concept* for convenience. MMPB includes 111 such concepts, spanning four categories: human, animal, object, and character. The *human* category is associated with both recognition and preference tasks, while the remaining categories are used solely for recognition tasks.

Queries will be constructed based on the following three task types. Since the required images differ depending on the task type, please refer to the descriptions below when collecting images. The three task types are as follows:

- **Awareness**: Can the model accurately identify *concept* in a given image?
- **Appropriateness**: Does the model activate *concept* only when it is contextually appropriate?
- **Coherency**: Does the model contradict *concept* in its responses?

### Image Collection

**Please first refer to the image provided for concept injection, and ensure that the images you collect are not identical to the injection image under any circumstances. The collected images must differ from the injection image to avoid data leakage and ensure proper evaluation.**

To evaluate **Awareness**, we will collect **positive images**—that is, images containing the same entity as the given concept. Since MMPB assumes static appearance and preference for each concept, all collected images should reflect a consistent visual identity and taste.

When collecting images of people, please avoid photos where the individual is wearing items that may obstruct recognition, such as sunglasses, hats, or other occluding accessories.

Below is an examples of correctly (left) and incorrectly (right) collected image:

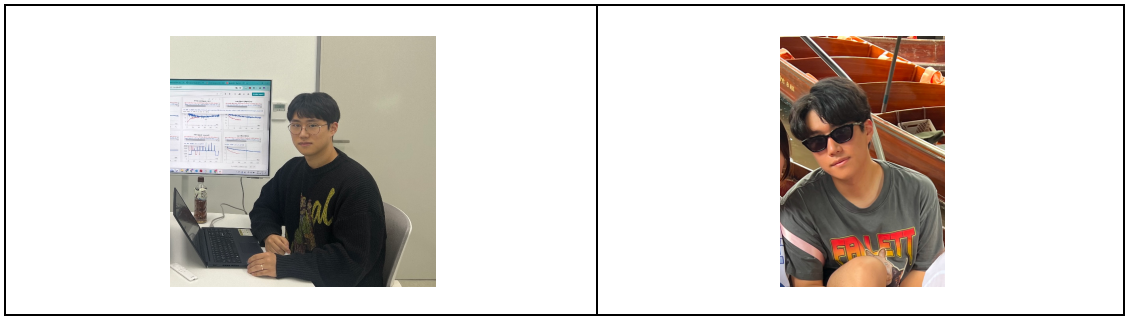

In the example above, although both images depict the same person, the one on the right—with sunglasses—makes it difficult to recognize the entity.

Regarding **preference**, please refer to the provided preference keywords for the given individual and collect images that align with those preferences.

Below is an example of properly collected image (right) that aligns with a preference (left).

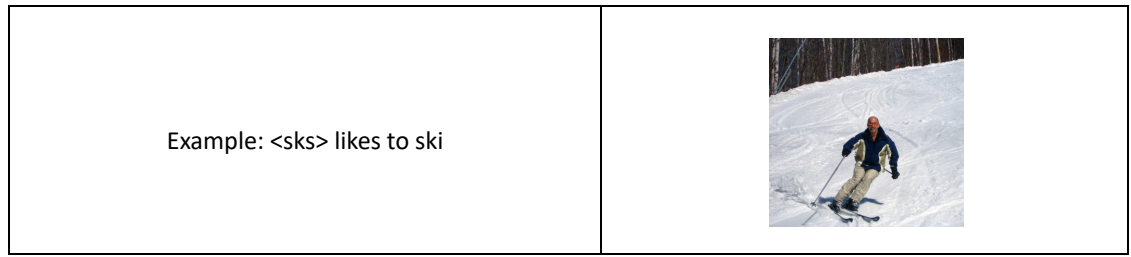

Example: <sks> likes to ski

For **Appropriateness**, please collect **negative images**—that is, images that do **not** contain the target concept. In the **animal** category, make sure to include both *same-species* and *different-species* examples. This distinction will allow us to later evaluate model performance under more subtle or confusing conditions.

You may also reuse images collected for Awareness (*i.e.,* positive examples for one concept) as negative examples for other concepts where appropriate. For **preference-related tasks**, collect images that do **not** align with the individual's preference keywords. For instance, if the concept is associated with "volcano travel" or "nail art," a mismatched image would be something unrelated like a hardware store or a sports event. Suggested disjoint keywords include: **Volcano Travel**, **Nail Art**, **Home Interior Design**, etc.

For **Coherency**, we will later generate multiple-choice questions using **positive images** only. Therefore, please collect **additional positive images** of the concept or reuse those from the Awareness set. For human preference concepts, we recommend including images of places or objects like **movie theaters**, **TV remotes**, or **shopping malls**, as these will support richer, contextually appropriate question generation later.

### Image Source

The following are the recommended sources for collecting images. For the **human** category, as mentioned earlier, consistent appearance across images is essential—we strongly recommend using **MC-LLaVA** for this purpose. For **character** and **object** concepts, high-quality, well-organized images are available in **MC-LLaVA** and **MyVLM**, respectively, and we strongly encourage leveraging both datasets.

- **MC-LLaVA** (located at /workspace/MC-LLaVA/)
- **MyVLM** (located at /workspace/MyVLM/)
- **Flickr** (https://www.flickr.com/ ): Please enable *All Creative Commons* license when collecting images.
- **Pexels** (https://www.pexels.com/ ): All images on this platform are free to use.
- **Google Image Search**: Use with a *Creative Commons* filter if possible. If no such filter is available, please include a comment with the **source URL** of the image.



\subsection{Query Generation}

We present the guidelines used by both models and human annotators during the query generation process.

\subsubsection{Models}

\begin{tcolorbox}[colback=gray!10, colframe=black!80, boxrule=0.5pt, arc=2pt, title=Guideline for model-based recognition query generation]
You will be given: 1) Task type 2) An image 3) A description of target concept <sks>\\

\textbf{Task:} Generate candidate VQA queries of three types—Awareness, Appropriateness, and Coherency.

\medskip
\textbf{Awareness}: Formulate both affirmative and negative yes/no questions about whether <sks> is present in the image, e.g.: Is <sks> not present in the image?
, Can you see <sks> in the photo?

\medskip
\textbf{Appropriateness} : Similar to Awareness, formulate both affirmative and negative yes/no questions about whether <sks> is present in the image.

\medskip
\textbf{Coherency}: Create one 4-option multiple-choice question where:
    \begin{itemize}
      \item Correct answer reflects the true relationship of <sks> to the image.
      \item Distractors are visually plausible but incorrect without concept reasoning.
      \item For single-entity images, distractors may consist exclusively of plausible options.
    \end{itemize}

\textbf{Always:} Use clear grammar and concise phrasing.
\end{tcolorbox}

\begin{tcolorbox}[colback=gray!10, colframe=black!80, boxrule=0.5pt, arc=2pt, title=Guideline for model-based preference query generation]
You will be given: 1)Task type 2) An image 3) A preference description of target concept <sks>\\

\textbf{Task:} Generate candidate VQA queries of three types—Awareness, Appropriateness, and Coherency—while filtering out text-only or image-only solvable questions for coherency.

\medskip
\textbf{Awareness}: Formulate both affirmative and negative yes/no questions about whether the image is related to <sks>’s preference, e.g.: Is the image tied to <sks>’s positive or negative liking?, Is the image not tied to <sks>’s positive or negative preference?

\medskip
\textbf{Appropriateness} : Similar to Awareness, formulate both affirmative and negative yes/no questions about whether the image is related to <sks>'s preference.

\medskip
\textbf{Coherency}: Create one 4-option multiple-choice question where:
    \begin{itemize}
      \item Correct answer reflects the true relationship of <sks> to the image.
      \item Distractor B is concept-aligned with <sks> but does not match the image.
      \item Distractors C and D are visually plausible but incorrect without concept reasoning.
    \end{itemize}
\textbf{Example}: \\
    \begin{verbatim}
Q Among the activities that could reasonably occur
in the given image, which one is <sks> least likely to be doing?
A) <correct>
B) <concept-aligned but not in image>
C) <visually plausible but concept-wrong>
D) <visually plausible but concept-wrong>
    \end{verbatim}
\textbf{Always:} Use clear grammar and concise phrasing.
\end{tcolorbox}

\includepdf[
scale=0.90,pages=4,pagecommand=\subsubsection{Humans}\label{sup:guide_query_gen_human}]{figures/guides/MMPB_tot_guidelines.pdf}

\newpage
\subsection{Query Evaluation}

After drafting each query, we send the concept description (appearance and, for humans, preferences), the image, and the draft query with its answer to three models (\S\ref{sec:construct}). If at least two models flag the same issue, we forward that flag to human annotators, who then apply our guidelines to revise the query. This model‐assisted flagging speeds up human quality checks.

\subsubsection{Models}
\begin{tcolorbox}[colback=gray!10, colframe=black!80, boxrule=0.5pt, arc=2pt, title=Flagging guidelines for models]

You will be provided with a personalizable concept’s textual descriptions, covering appearance and preferences, and a set of image-query-answer pairs about an image. Your task is to identify whether each pair exhibits any of the following issues:

\vspace{0.5em}
\textbf{1. Trivial} \par
A query is \textbf{Trivial} if it can be answered directly from the query itself or by using commonsense, without needing the image or any extra information.

\textit{Example:} \par
Q: What is \textless sks\textgreater{} doing in the movie theater? \\
A: Watching a movie. \\
Judgement: Trivial \\
(The query already states “movie theater,” and watching a movie in a movie theater is trivial.)

\vspace{1em}
\textbf{2. Concept-Only} \par
A query is \textbf{Concept-Only} if it can be answered using only the provided textual description (appearance, preferences, etc.), without looking at the image.

\textit{Example:} \par
Q: What is \textless sks\textgreater{}’s gender? \\
A: Male. \\
Judgement: Concept-Only \\
(The description explicitly says \textless sks\textgreater{} is male, so there’s no need to inspect the picture.)

\textit{Example:} \par
Q: What is \textless sks\textgreater{} likely to be doing here? Choice: Playing basketball / Yoga / Drinking coffee / Ice-fishing \\
A: Playing basketball. \\
Judgement: Concept-Only \\
(The description explicitly says \textless sks\textgreater{} likes to play basketball, so there’s no need to inspect the picture.)

\vspace{1em}
\textbf{3. Image-Only} \par
A query is \textbf{Image-Only} if it can be answered using only the provided image, without needing the personal description.

\textit{Example:} \par
Q: (An image with a man riding a bike) What is \textless sks\textgreater{} who is wearing a jacket doing? \\
A: Riding a bike. \\
Judgement: Image-Only \\
(The query describes \textless sks\textgreater{} in the image, so there’s no need to inspect the personal description.)

\textit{Example:} \par
Q: (An image with a man playing basketball) What is \textless sks\textgreater{} likely to be doing here? Choice: Playing basketball / Yoga / Drinking coffee / Ice-fishing \\
A: Playing basketball. \\
Judgement: Image-Only \\
(If the answer can be determined from the image alone and all other choices are unrelated to it, then display only the image.)

\vspace{1em}
Use exactly these three labels—\textbf{Trivial}, \textbf{Concept-Only}, and \textbf{Image-Only}—when annotating. Refer back to the definitions and examples above.

\end{tcolorbox}
\newpage

\subsubsection{Humans}
\label{supsec:human_guide_eval}
\begin{tcolorbox}[colback=gray!10, colframe=black!80, boxrule=0.5pt, arc=2pt, title=Query evaluation guidelines for human annotators]
You will be provided with:
\begin{itemize}
  \item \textbf{Model issue flagging}
  \item \textbf{A personalizable concept’s textual descriptions}, covering appearance and preferences
  \item \textbf{A set of image–query–answer pairs} about a given image
\end{itemize}

\texttt{<Guideline for models>}\\

\textbf{1.} Carefully review the issue the model has flagged.\\
\quad $\bullet$ If the flagged issue is valid, correct the answer choices accordingly.\\
\quad $\bullet$ If the flagged issue is invalid but there is a different valid issue, correct the answer choices accordingly.\\
\quad $\bullet$ If the flagged issue is invalid and there is no other issue, annotate \texttt{None} above the question.\\[6pt]

\textbf{2.} Ensure every question is phrased as an open-ended, free-form query. \\
\quad $\bullet$ If a question is not open-ended, rewrite it so it can be answered freely.\\[3pt]

\emph{Example of an open-ended question:}\\
Q. What is \sks{} doing in the movie theater?\\[6pt]

\emph{Example of a choice-based question (to be avoided):}\\
Q. Which of the following four options best describes what \sks{} might be doing?
\end{tcolorbox}

\section{Human Evaluation Platforms}
\label{sup:humaneval}
\begin{figure}[ht]
    \centering
    \includegraphics[width=0.8\linewidth]{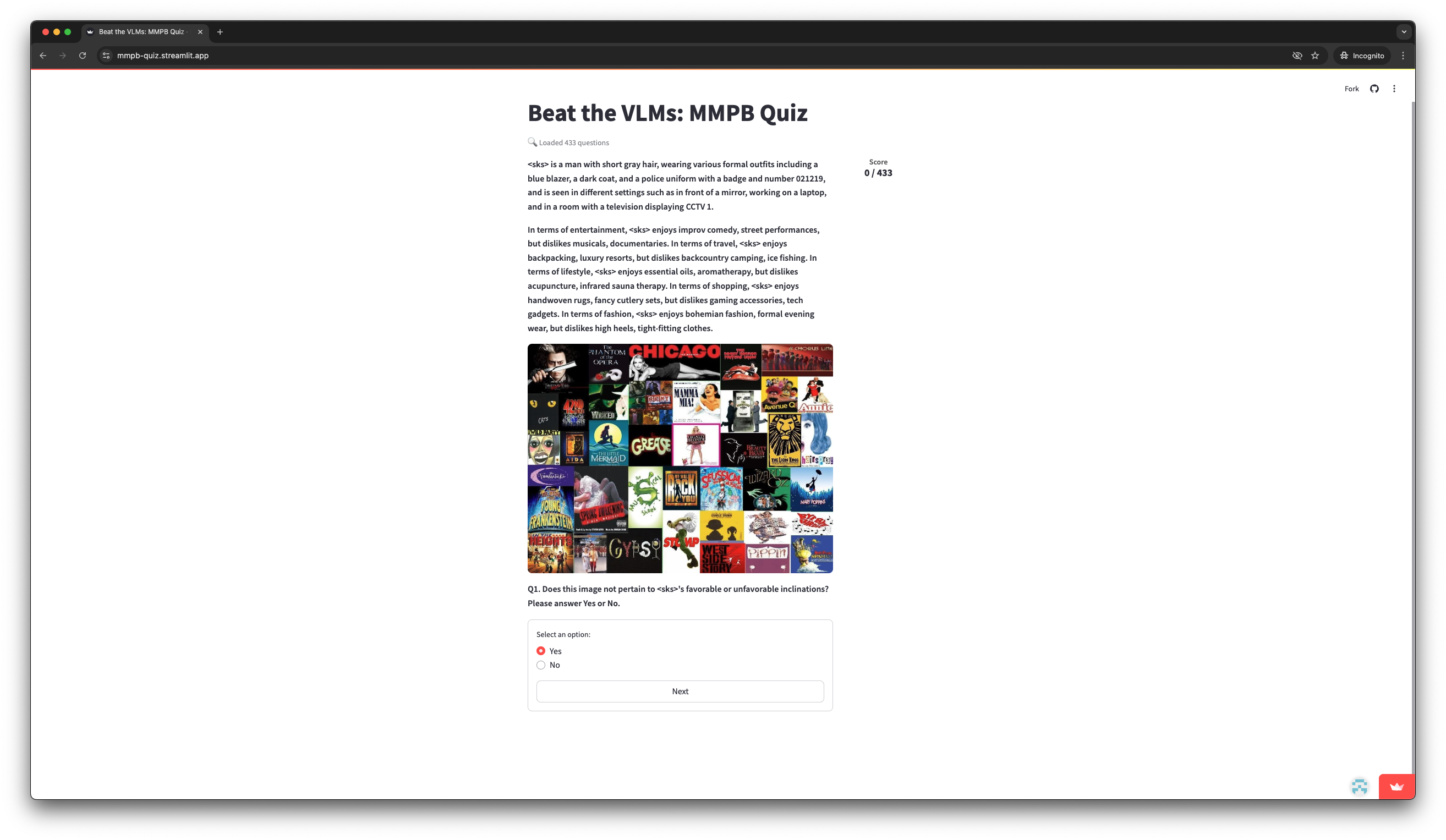}
\caption{Interface of the human evaluation platform.}
        \label{fig:platform_human_eval}
\end{figure}

\begin{figure}[htbp]
  \centering
  \begin{subfigure}[b]{0.32\linewidth}
    \centering
    \includegraphics[width=\linewidth]{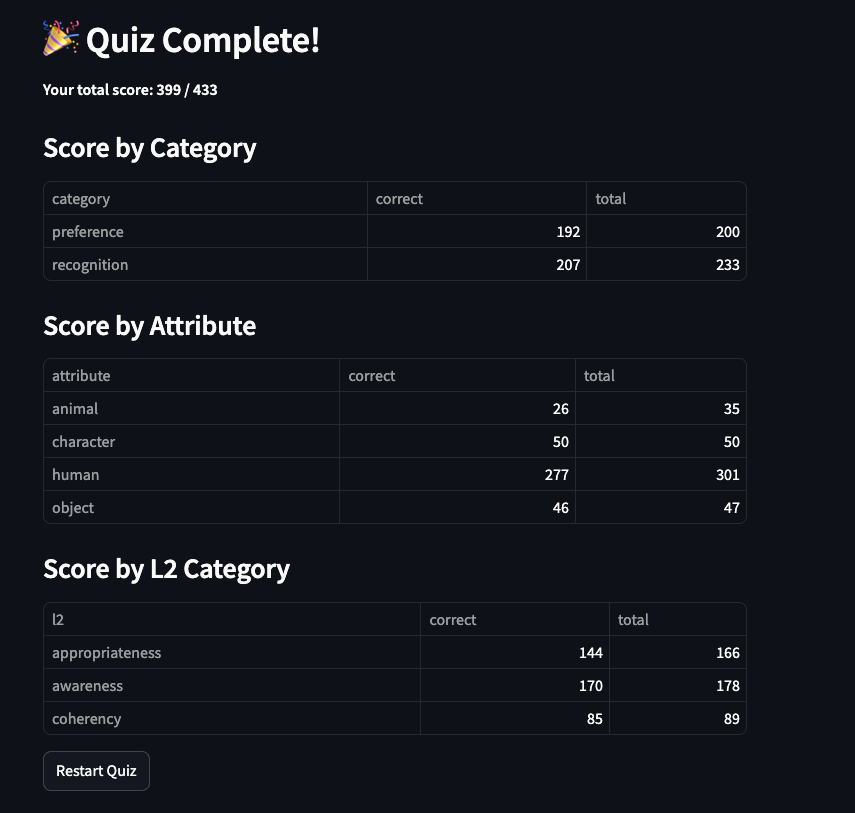}
    \caption{}
    \label{fig:ressub1}
  \end{subfigure}\hfill%
  \begin{subfigure}[b]{0.32\linewidth}
    \centering
    \includegraphics[width=\linewidth]{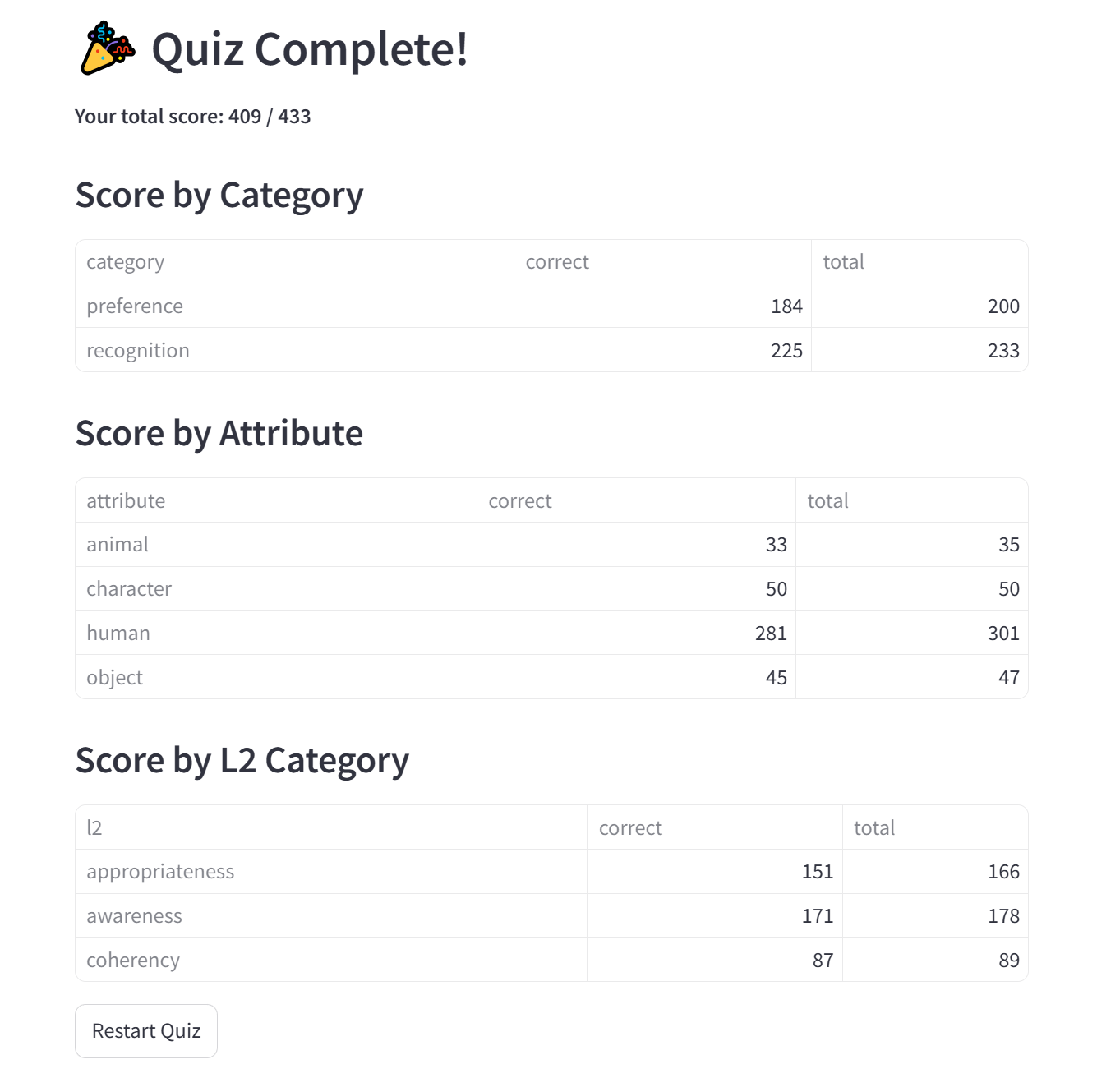}
    \caption{}
    \label{fig:ressub2}
  \end{subfigure}\hfill%
  \begin{subfigure}[b]{0.32\linewidth}
    \centering
    \includegraphics[width=\linewidth]{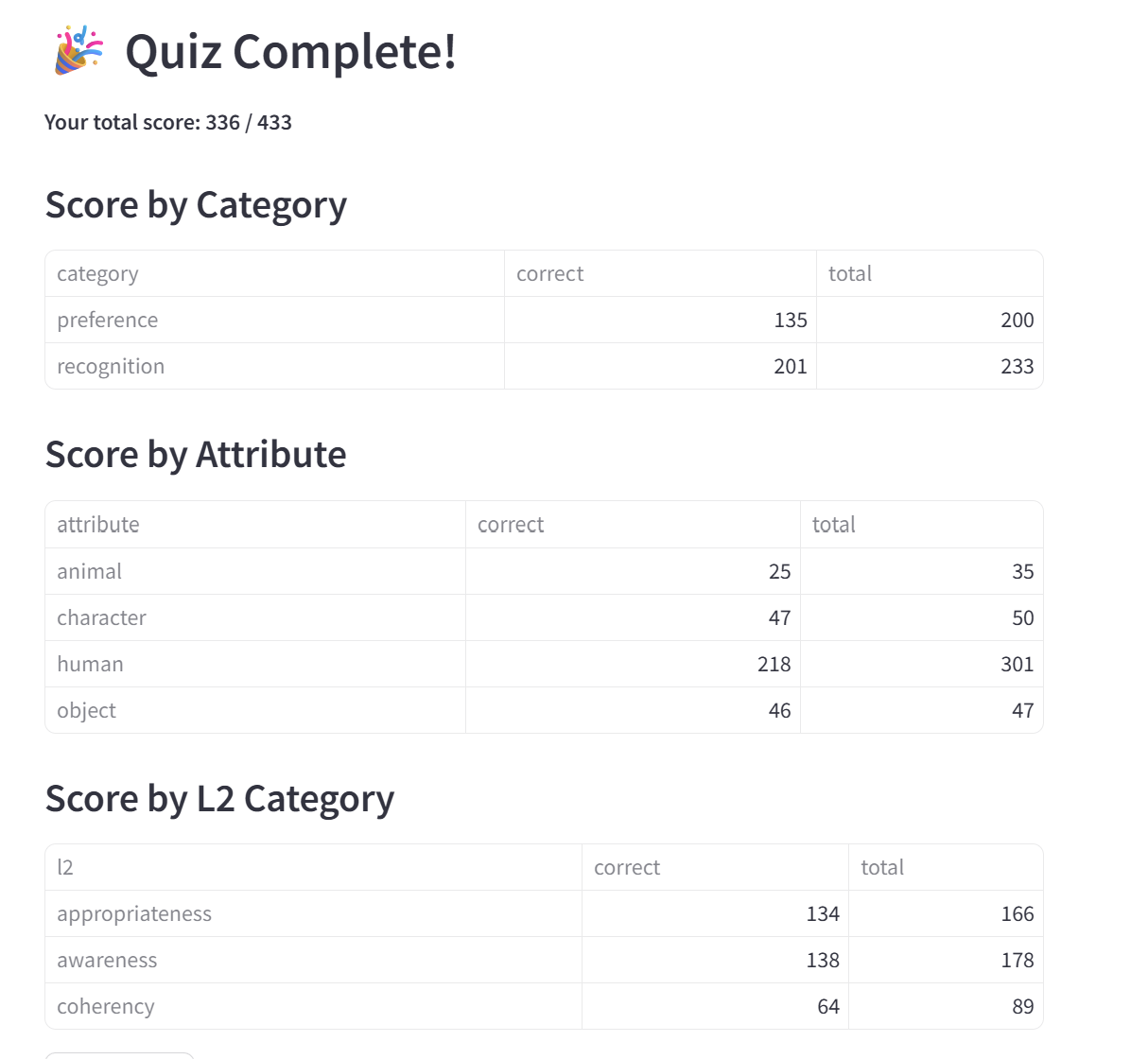}
    \caption{}
    \label{fig:ressub3}
  \end{subfigure}

  \caption{Examples of the human evaluation results interface.}
  \label{sup:human_eval_results}
\end{figure}

Human baseline evaluation was conducted via a Streamlit\footnote{https://streamlit.io/} platform. Evaluators manually assessed queries drawn from the \textit{Moderate} concept‐injection set. Since \mmpb{} comprises 10,017 total questions, asking humans to label them all would be prohibitively time‐consuming and risk fatigue‐related bias. To ensure a fair yet feasible evaluation, we randomly sampled one concept from each of five categories (two human concepts, one animal, one object, and one character), yielding a representative subset across concept types. Figure~\ref{fig:platform_human_eval} illustrates the evaluation interface. Once all assessments were complete, we computed scores by category and task type, then aggregated them to produce an overall human‐baseline accuracy (Figure~\ref{sup:human_eval_results}, see Figure~\ref{fig:main_results} for results). We plan to open‐source this evaluation platform in the near future, and will also release the five‐concept dataset used by human evaluators as the \mmpb{}-Mini version.

\section{More Information about \mmpb{}}
\label{appendix:info}

\begin{table}[ht]
  \centering
    \caption{Computational resources for evaluating \mmpb{}.}
  \begin{subtable}[t]{0.525\textwidth}
    \centering
    \begin{tabular}{lc}
      \toprule
      Family           & Cost \\ 
      \midrule
      GPT-4o           & \$224.5 \\ 
      Claude-Sonnet    & \$178.4 \\ 
      Gemini-Flash     & \$26.1  \\ 
      \bottomrule
    \end{tabular}
    \caption{Total API cost.}
    \label{sup:api_cost}
  \end{subtable}\hfill
  \begin{subtable}[t]{0.45\textwidth}
    \centering
      \renewcommand{\arraystretch}{0.7}
    \begin{tabular}{lc}
      \toprule
      Method        & Avg. Runtime (h) \\ 
      \midrule
      Text 0 Turn   & 2.2 \\ 
      Text 10 Turn  & 2.7 \\ 
      Image 0 Turn  & 4.2 \\ 
      Image 10 Turn & 4.5 \\ 
      \bottomrule
    \end{tabular}
    \caption{Average runtime.}
    \label{sup:runtime}
  \end{subtable}
  \label{sup:cost_and_runtime}
\end{table}

\subsection{Computational Resources}
Open-source models were run locally on four NVIDIA H100 GPUs. All models can be evaluated on a single GPU, except for InternVL2.5-78B, Qwen2-VL-72B, Qwen2.5-VL-72B, and LLaVA-OV-72B, which were evaluated using three GPUs.

Closed-source models were evaluated via their APIs using batch processing. Detailed total costs for each model family are provided in Table~\ref{sup:api_cost}.

Table~\ref{sup:runtime} reports the average runtime of all open-source models used in benchmarking. The \textit{Text} method refers to \textit{Moderate} description-based concept injection, whereas the \textit{Image} method refers to \textit{2-image} concept injection.

\subsection{Statistics}
\label{sup:statistics}
\begin{figure}[h]
\centering
    \includegraphics[width=0.65\linewidth]{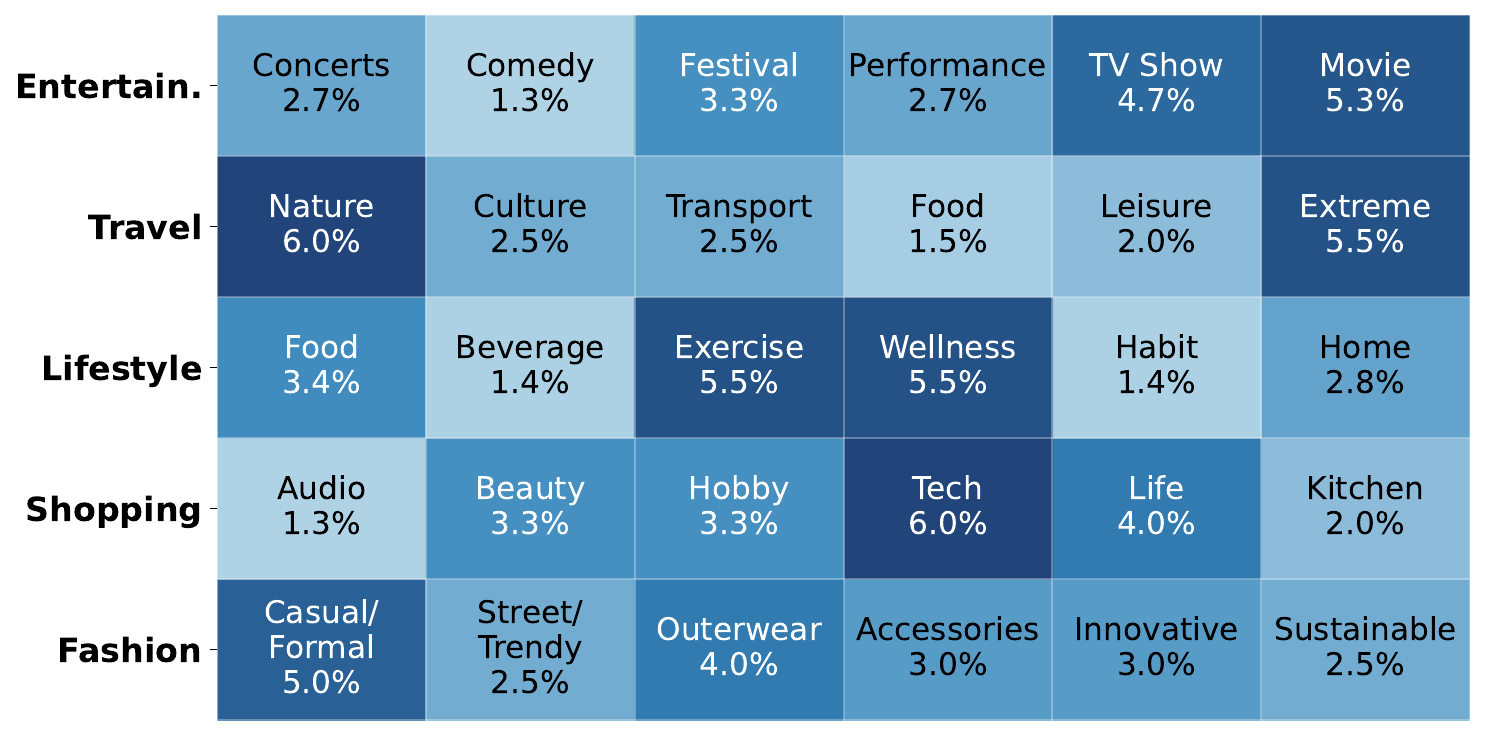}
\caption{Distribution of preference keywords.}
    \label{fig:dataset_statistics}
\end{figure}

\begin{figure}[!h]
  \centering
  \begin{subfigure}[t]{0.1955\textwidth}
    \includegraphics[width=\linewidth]{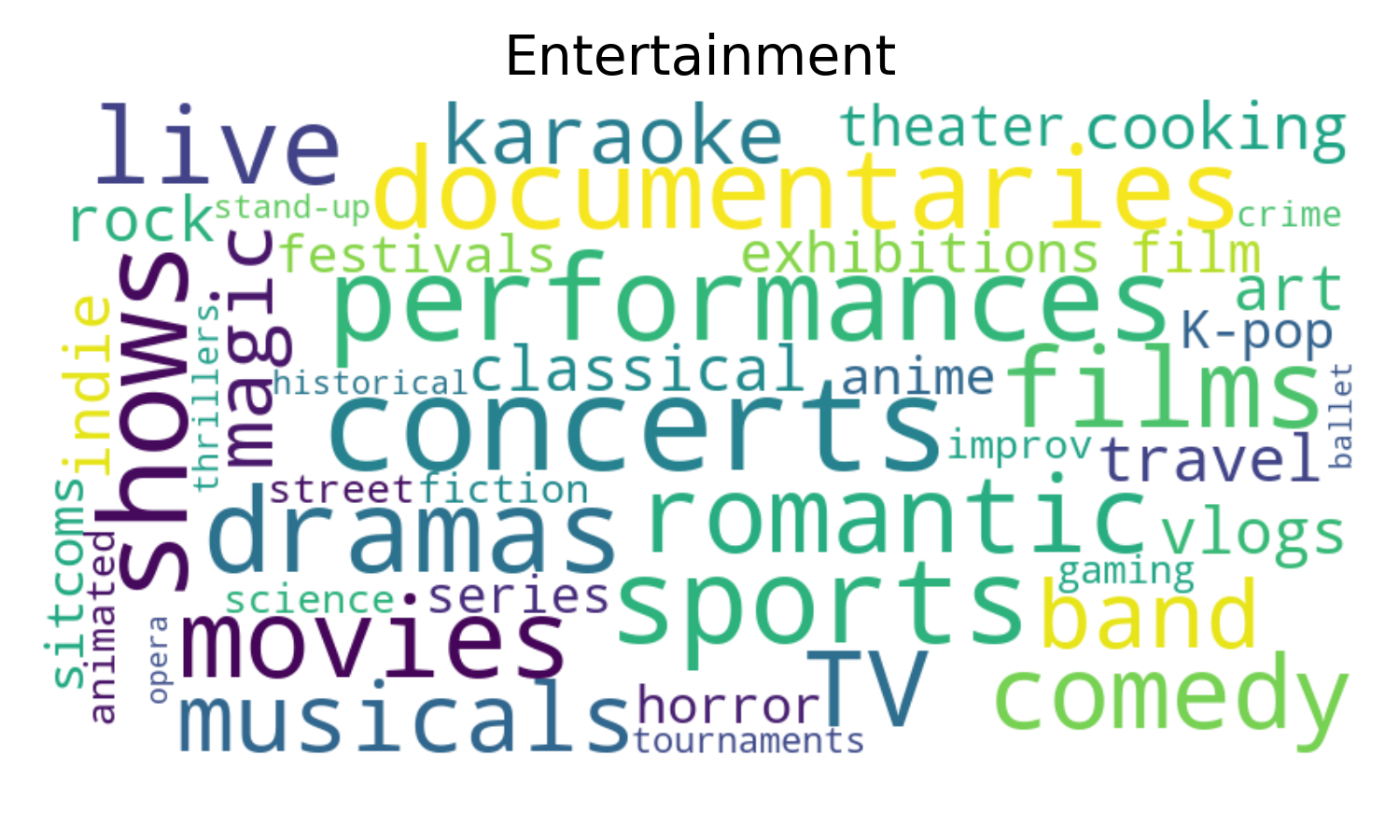}
  \end{subfigure}
  \hfill
  \begin{subfigure}[t]{0.19525\textwidth}
    \includegraphics[width=\linewidth]{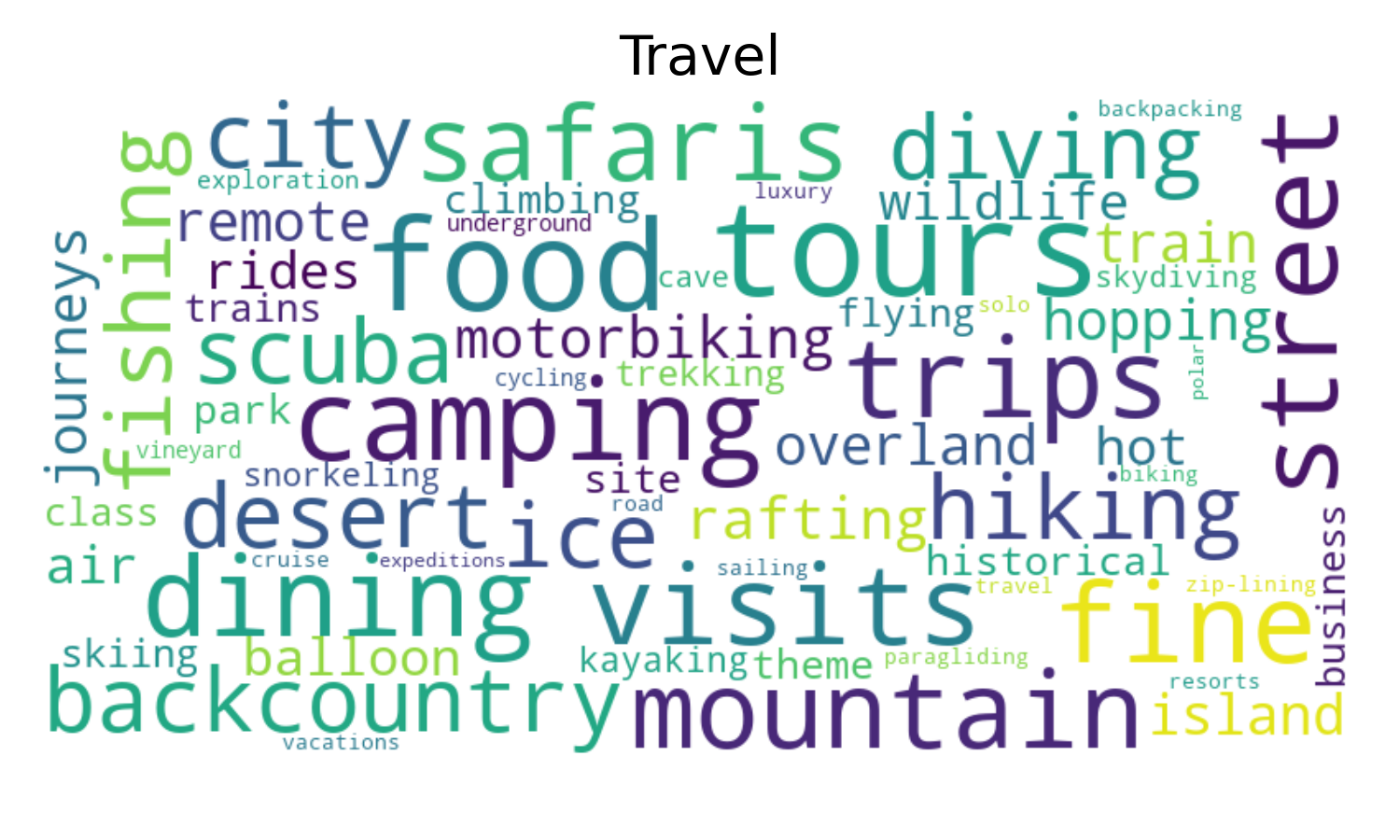}
  \end{subfigure}
  \hfill
  \begin{subfigure}[t]{0.19525\textwidth}
    \includegraphics[width=\linewidth]{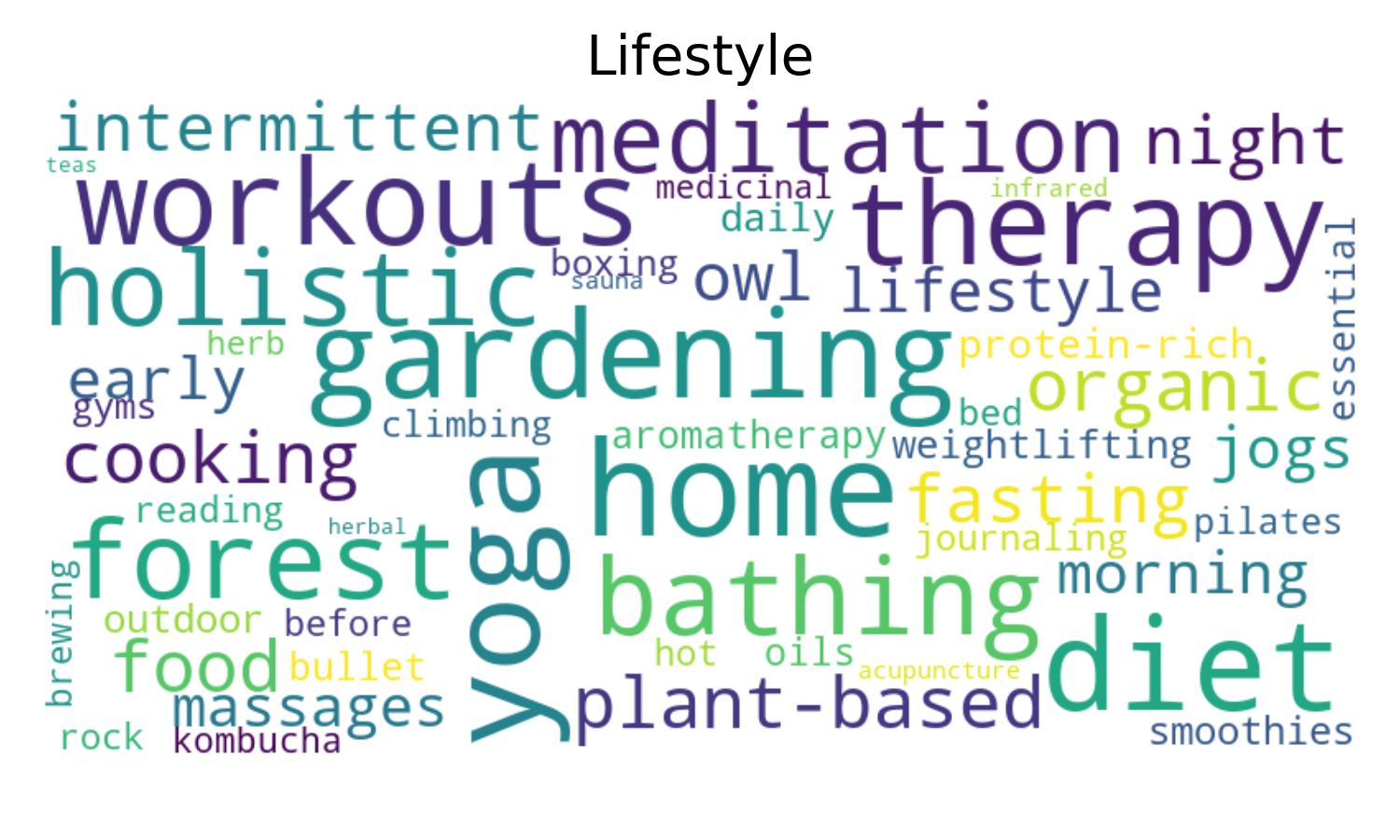}
  \end{subfigure}
  \hfill
  \begin{subfigure}[t]{0.19525\textwidth}
    \includegraphics[width=\linewidth]{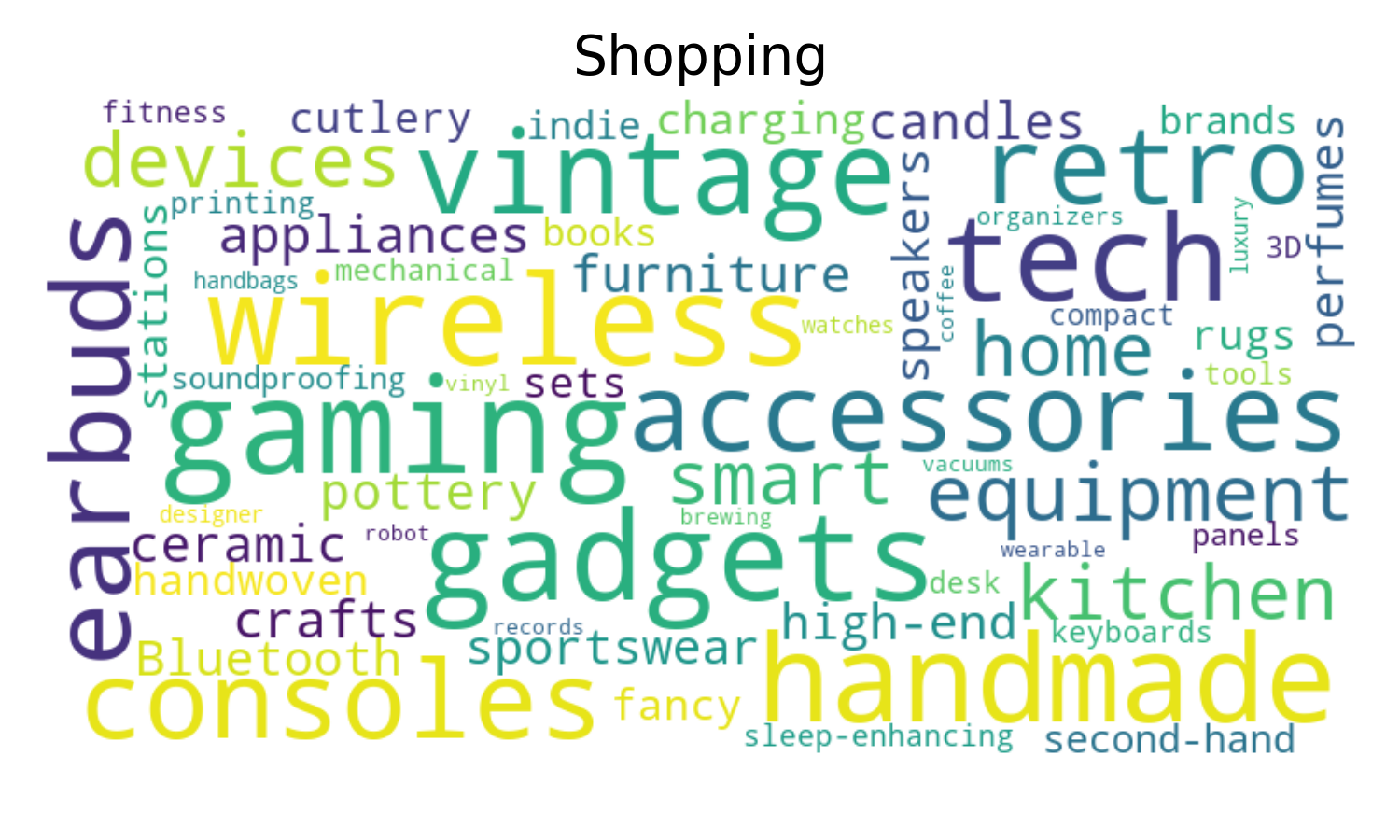}
  \end{subfigure}
  \hfill
  \begin{subfigure}[t]{0.19525\textwidth}
    \includegraphics[width=\linewidth]{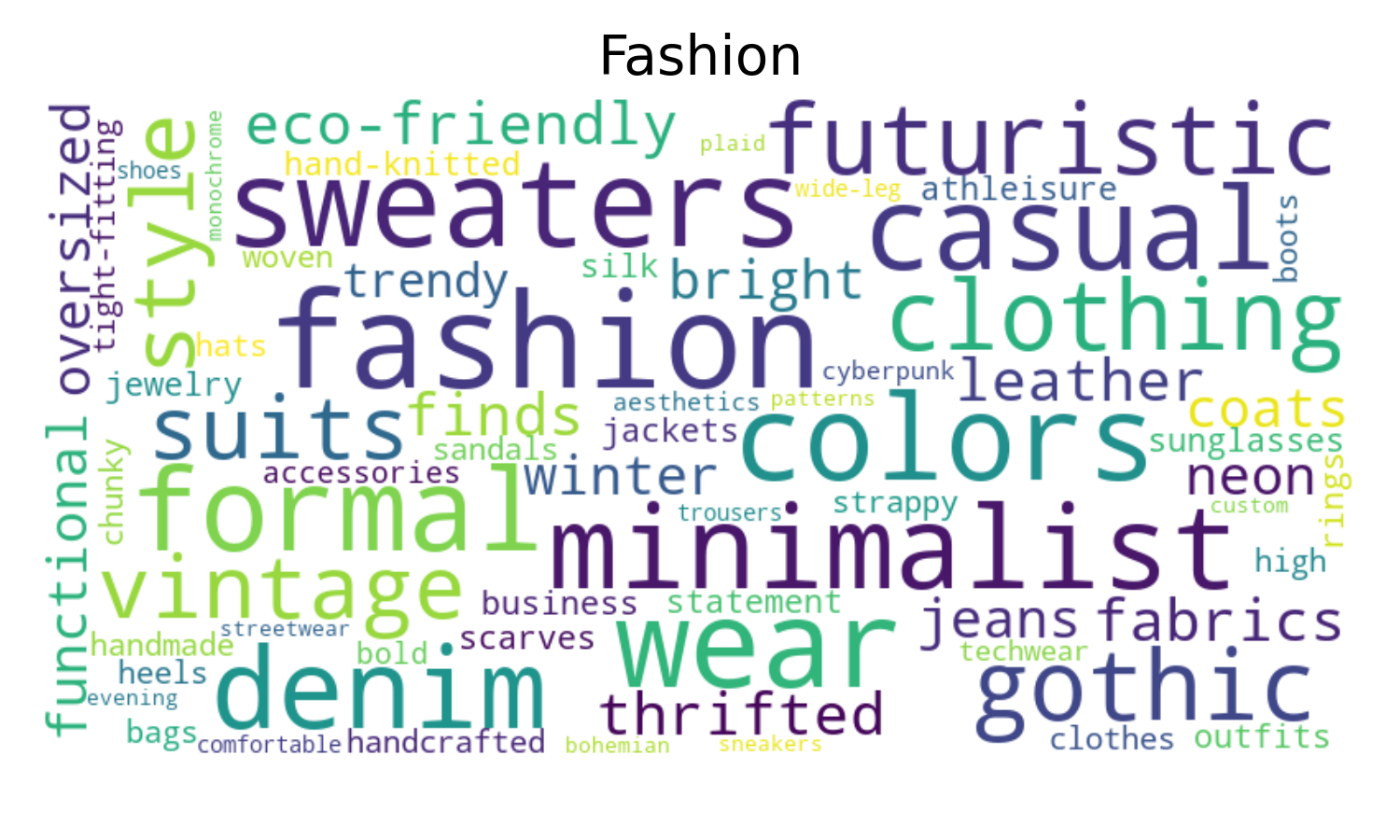}
  \end{subfigure}

\caption{Word clouds of preference keywords in \mmpb{} across five domains.}
  \label{sup:wc_pref}
\end{figure}

Table~\ref{suptab:statistics} presents overall statistics for \mmpb{}, including the number of concepts and the question counts for each task type; the dataset comprises 10,017 image–text query pairs in total. As shown in Figure~\ref{fig:dataset_statistics}, human preference keywords are drawn from 30 subdomains across five domains: Entertainment, Travel, Lifestyle, Shopping, and Fashion. Figure~\ref{sup:wc_pref} displays word clouds of the extracted keywords for each domain, generated by GPT-4o.

\begin{table}[h]
\centering
\caption{\mmpb{} statistics, including the number of concepts and question counts across task types.}
\begin{tabular}{lcccccc}
\toprule
Category      & \# Concept & Awareness & Appropriateness & Coherency & Total \\
\midrule
Human (Rec.)   &  50  &   734  &           1400  &      367  &  2501 \\
Human (Pref.)  &  50 &    2500  &           1250  &     1250  &  5000 \\
Animal         &   20 &   200  &            400  &      100  &   700 \\
Object         &   25 &    344  &            500  &      172  &  1016 \\
Character      &   16 &    320  &            320  &      160  &   800 \\
\midrule
- & - & - & -& - & 10017\\
\bottomrule
\end{tabular}
\label{suptab:statistics}
\end{table}

\newpage




\section{Additional Results}
\label{appendix:add_results}

\subsection{Personalization Fine-tuning}
\label{sup:pfinetuning}

We additionally fine-tuned LLaVA-1.5-13B with two strategies: soft prompt tuning and personalized LoRA. 
As shown in Table~\ref{suptab:finetune}, both methods yield clear improvements over the baseline, 
with LoRA achieving the strongest gains. 
These results confirm that \mmpb{} reliably captures personalization effectiveness.

\begin{table}[h]
\centering
\caption{Performance of LLaVA-1.5-13B on \mmpb{} with different personalization methods.}
\begin{tabular}{lcccc}
\toprule
Method          & Epoch & Preference & Recognition & Overall \\
\midrule
Baseline        &   -   &   47.8     &   59.6      &   51.5  \\
Soft Prompt     &  10   &   57.8     &   59.6      &   58.4  \\
Soft Prompt     &  30   &   58.4     &   66.5      &   61.0  \\
Soft Prompt     &  50   &   57.6     &   71.3      &   62.1  \\
LoRA            &  10   &   52.0     &   72.1      &   58.0  \\
LoRA            &  30   &   58.7     &   86.1      &   67.1  \\
LoRA            &  50   &   60.0     &   58.1      &   59.4  \\
\bottomrule
\end{tabular}
\label{suptab:finetune}
\end{table}

\subsection{Model Sizes}
\paragraph{Impact of model scale on \textit{Persistency}}

\begin{figure}[t]
  \centering
  \begin{subfigure}[t]{0.475\textwidth}
    \includegraphics[width=\linewidth]{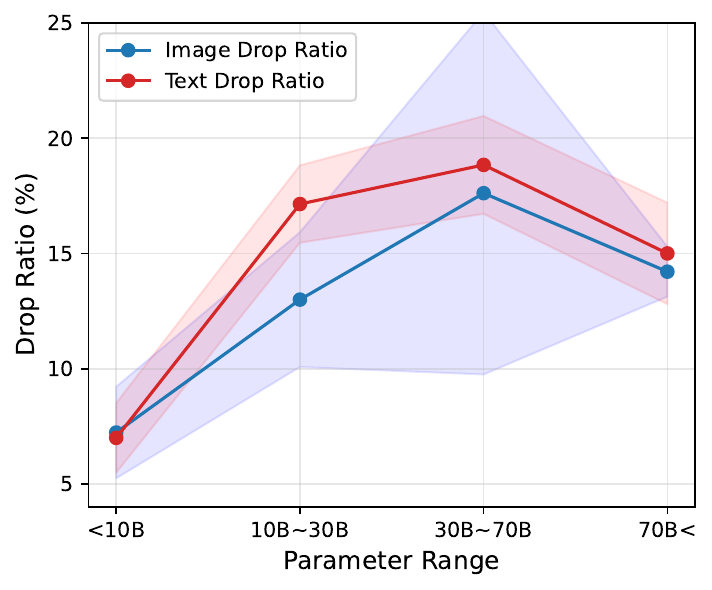}
  \end{subfigure}
  \hfill
  \begin{subfigure}[t]{0.475\textwidth}
    \includegraphics[width=\linewidth]{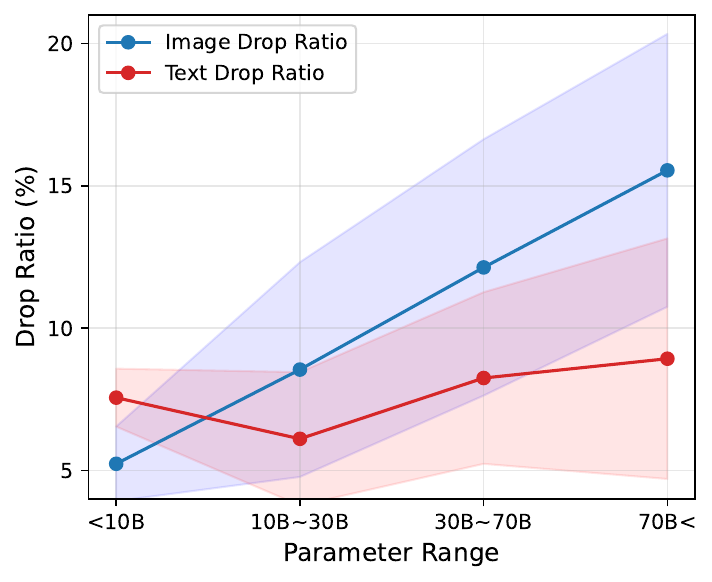}
  \end{subfigure}
    \caption{Multi-turn conversation performance drop ratios by injection modality. (left) preference accuracy drop; (right) recognition accuracy drop.}
\label{supfig:param_drop}
\end{figure}

We analyze multi‐turn conversation drop ratios for both injection modalities (Figure~\ref{supfig:param_drop}). For preference tasks, both text‐ and image‐based injections show similar drop trends as turns increase. This occurs because, in both modalities, preference injection is carried out through the text modality. However, in recognition tasks, image‐based injection yields an almost linear performance decline, whereas text‐based injection maintains far more stable drop rates. In other words, larger models struggle to retain personalized context when it is delivered via images. This result reinforces the suboptimal nature of image‐based personalization discussed in \S\ref{subsec:challenges}.

\begin{table}[h]
\centering
\caption{Ovis2-34B performance using human-authored vs. model-generated descriptions to test for self-advantage}
\begin{tabular}{lcc}
\toprule
Description      & 0 Turn & 10 Turn \\
\midrule
Self-Generated  &  73.3   & 63.6 \\
Human-created  &  75.3  & 63.6\\
\bottomrule
\end{tabular}
\label{suptab:ovis2}
\end{table}

\subsection{Comparison of Ovis2-34B with Human-Created Descriptions}
\label{supsubsec:ovis2}
In \S\ref{sec:construct}, we used Ovis2-34B to generate descriptions for all concept categories and then evaluated the model on MMPB (\S\ref{sec:benchmark}). To rule out any unfair advantage from using its own outputs, we asked human annotators to write descriptions from scratch (without seeing the model’s versions) for a held-out set of 10 concepts (4 human, 2 animal, 2 object, and 2 character). We then compared Ovis2-34B’s performance when using human-authored descriptions versus its own. As shown in Table~\ref{suptab:ovis2}, Ovis2-34B attains higher accuracy using the human-created descriptions than its own outputs. If our benchmark favored model-generated text, Ovis2-34B would perform best with its own descriptions. That Ovis2-34B performs better on independent, human-authored descriptions demonstrates that our evaluation does not unduly advantage self-generated content.

\subsection{Blind Test}
\label{sup:blind}
To evaluate the query quality of \mmpb{}, we conduct a blind test on six models (see \S\ref{subsubsec:suboptimal_visual_concept_injection} and \S\ref{subsubsec:multi_turn_degradation}). We \textbf{black out} the question images and provided only the text to each model, using the \textit{Moderate} text-based injection with 0-turn setting. This experiment tests whether the models produce predictable results when solving questions based solely on text. 

Ideally, since the models cannot see the images, they should always deny the concept in \textit{Awareness} tasks on positive images, resulting in 0\% accuracy. In \textit{Appropriateness} tasks, they should always agree, yielding 100\% accuracy. Finally, for \textit{Coherency}, our design ensures that two distractors are solvable only through visual information, while one is solvable only through the injected concept. Without the image, a correct concept injection implies a 50\% chance of selecting the correct answer—\ie a random choice between the true answer and the concept-only distractor.

As shown in Table~\ref{suptab:blind}, the blind-test results closely match these optimal values, once again demonstrating the effectiveness and robustness of our query design.  

\begin{table}
\centering
\small
\caption{Blind-test results for Recognition and Preference across task types.}
\begin{tabular}{l
  ccc  
  ccc  
}
\toprule
\multirow{2}{*}{Method} 
  & \multicolumn{3}{c}{Recognition} 
  & \multicolumn{3}{c}{Preference} \\
\cmidrule(lr){2-4} \cmidrule(lr){5-7}
  & Awareness & Appropriateness & Coherency 
  & Awareness & Appropriateness & Coherency \\
\midrule
Blind          & 0.0 & 100.0 & 41.5 
               & 15.4 & 82.0 & 41.4 \\
Optimal & 0.0 & 100.0 & 50.0 & 0.0 & 100.0 & 50.0 \\

\bottomrule
\end{tabular}
\label{suptab:blind}
\end{table}

\begin{figure}[h]
  \centering
  \begin{subfigure}[t]{0.495\textwidth}
    \includegraphics[width=\linewidth]{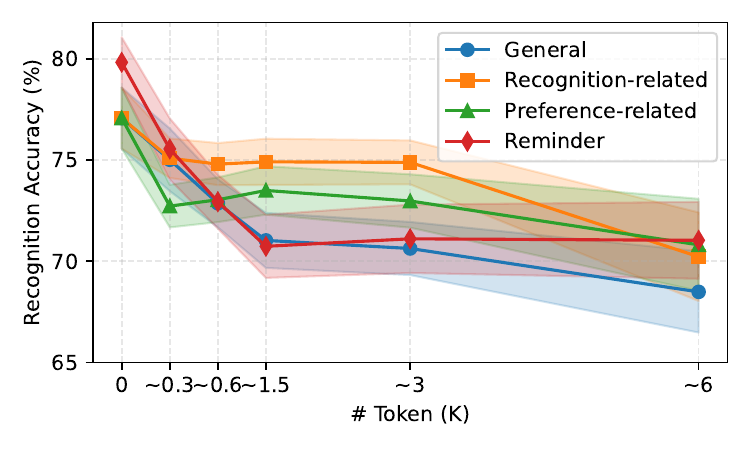}
  \end{subfigure}
  \hfill
  \begin{subfigure}[t]{0.495\textwidth}
    \includegraphics[width=\linewidth]{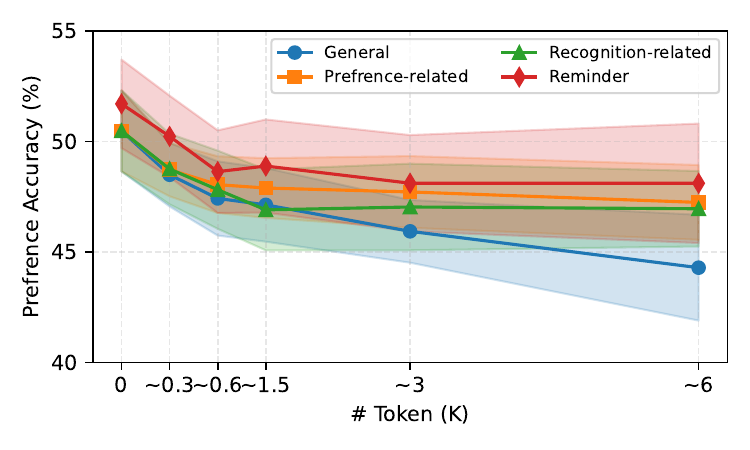}
  \end{subfigure}
\caption{Multi-turn conversation performance with \textit{Recognition-related} and \textit{Preference-related} topics: (left) recognition accuracy; (right) preference accuracy.}
    \label{supfig:diverse_mt}
\end{figure}

\subsection{Multi-turn Conversation}
\label{sup:add_mtturn}
We additionally compare VLM \textit{Persistency} under general conversation versus recognition‐ and preference‐related dialogues. We first crafted human‐written prompts for 10-turns on generic topics (\eg beard styling, hair dyeing, K-pop band recommendations, movie genres), ensuring they do not conflict with the concept’s preferences and maintain a neutral tone. We then used GPT-4o in a few-shot setup to augment multi-turn dialogues for each topic. Since dialogue lengths vary, we evaluate persistency based on cumulative token count rather than turn count. Example prompts for \textit{Recognition-Related} and \textit{Preference-Related} dialogues appear in Box~\ref{recognition_box} and Box~\ref{pref_box}, respectively. We also include the \textit{Reminder} prompting method for comparison.

As shown in Figure~\ref{supfig:diverse_mt}, general conversations incur the largest multi‐turn performance drop. \textit{Recognition‐related} dialogues yield the smallest drop in recognition accuracy, while for preference tasks all specialized dialogues (\textit{Recognition‐} or \textit{Preference‐related} and \textit{Reminder}) exhibit similar drop trends, saturating around 6k tokens. These results show that even general dialogues centered on recognition‐ or preference‐related topics deliver persistency benefits comparable to targeted reminder.

\subsection{Prompting Methods}
\label{supsec:prompting}
\begin{figure}[h] 
  \centering
  \captionsetup[subfigure]{justification=centering}

  \begin{subfigure}[b]{0.48\textwidth}
    \includegraphics[width=\linewidth]{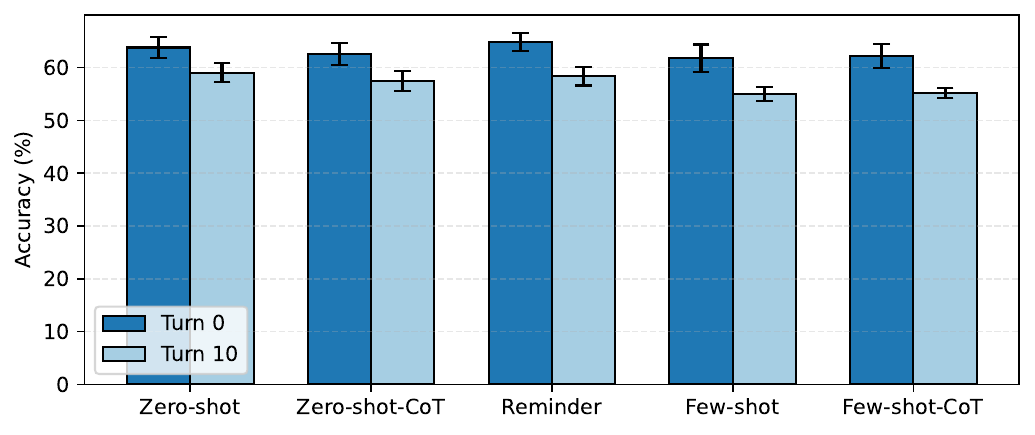}
    \caption{Overall}
    \label{fig:prompt1}
  \end{subfigure}\hfill
  \begin{subfigure}[b]{0.48\textwidth}
    \includegraphics[width=\linewidth]{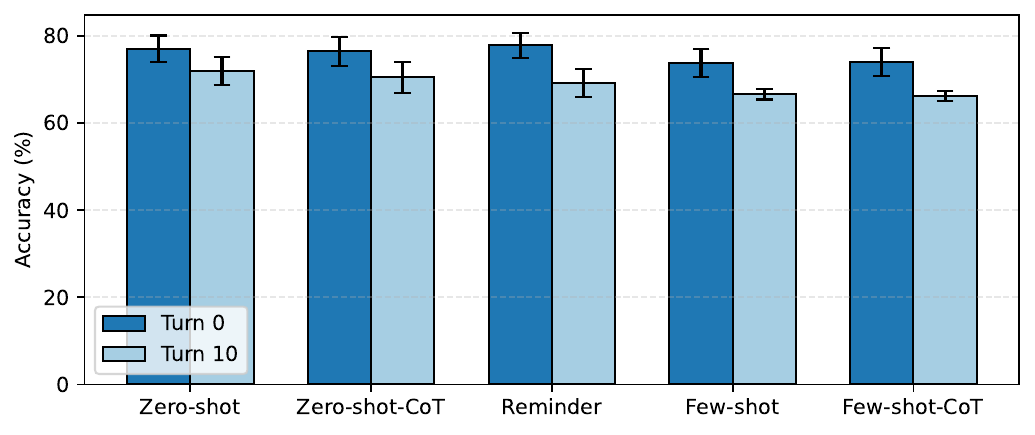}
    \caption{Recognition}
    \label{fig:prompt2}
  \end{subfigure}

  \vspace{0.5em}

  \begin{subfigure}[b]{0.49\textwidth}
    \includegraphics[width=\linewidth]{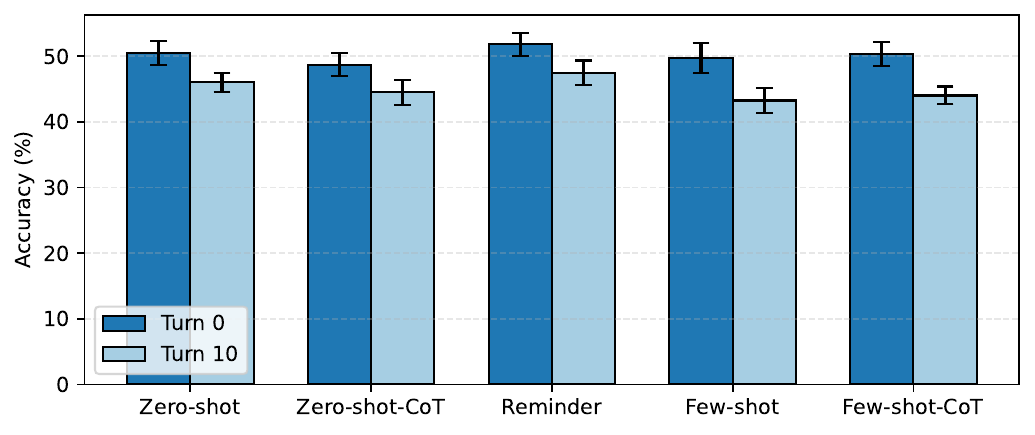}
    \caption{Preference}
    \label{fig:prompt3}
  \end{subfigure}\hfill
  \begin{subfigure}[b]{0.49\textwidth}
    \includegraphics[width=\linewidth]{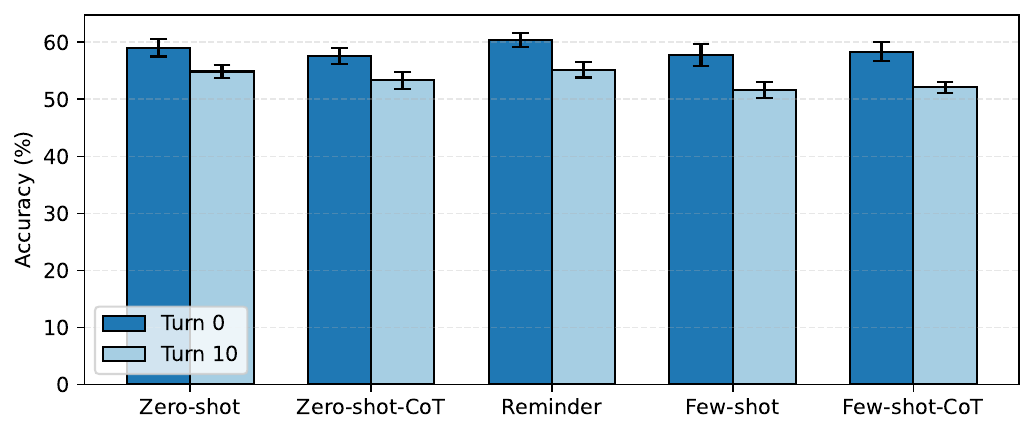}
    \caption{Human}
    \label{fig:prompt4}
  \end{subfigure}

  \vspace{0.5em}

  \begin{subfigure}[b]{0.49\textwidth}
    \includegraphics[width=\linewidth]{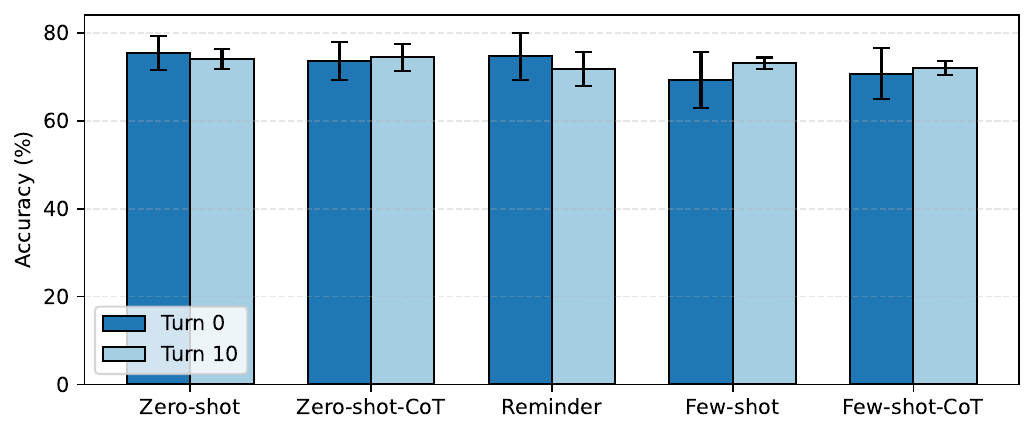}
    \caption{Animal}
    \label{fig:prompt5}
  \end{subfigure}\hfill
  \begin{subfigure}[b]{0.49\textwidth}
    \includegraphics[width=\linewidth]{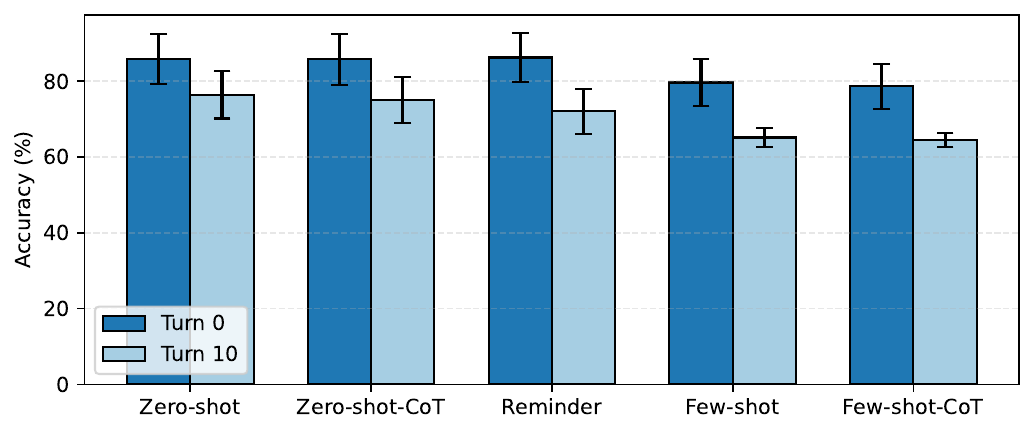}
    \caption{Object}
    \label{fig:prompt6}
  \end{subfigure}

  \vspace{0.5em}

  \begin{subfigure}[b]{0.49\textwidth}
    \includegraphics[width=\linewidth]{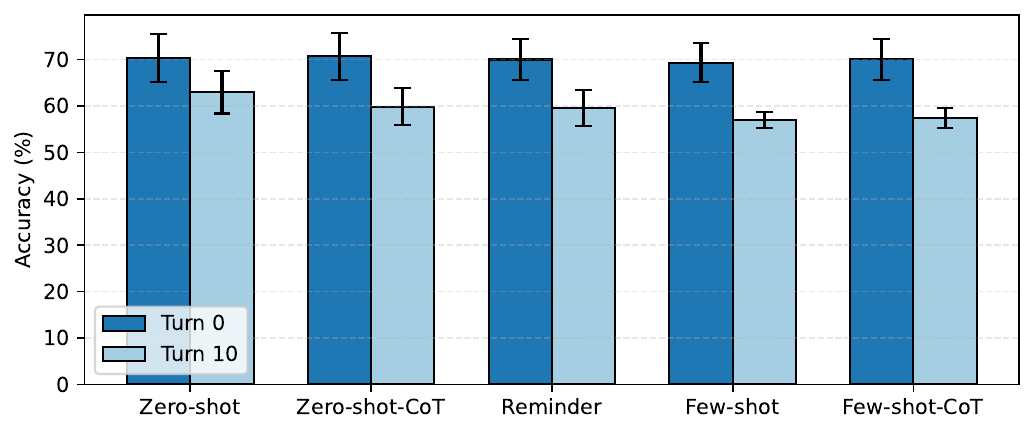}
    \caption{Character}
    \label{fig:prompt7}
  \end{subfigure}\hfill
  \begin{subfigure}[b]{0.49\textwidth}
    \includegraphics[width=\linewidth]{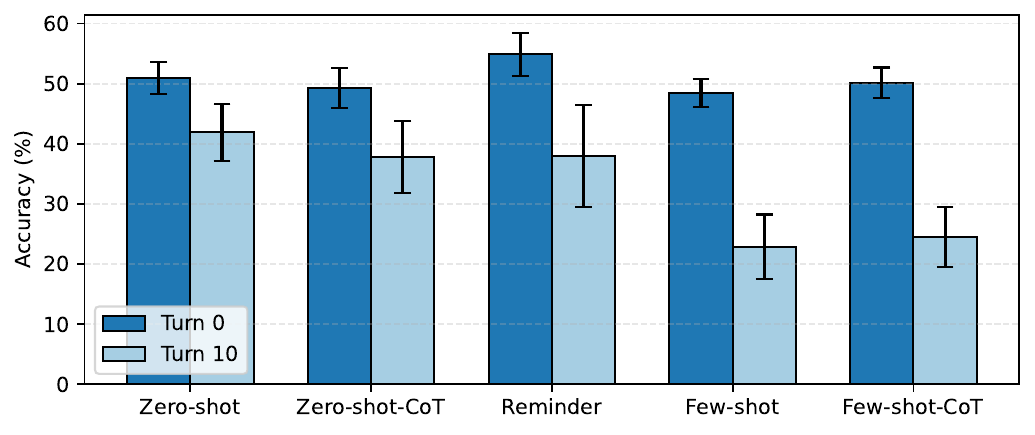}
    \caption{Awareness}
    \label{fig:prompt8}
  \end{subfigure}

  \vspace{0.5em}

  \begin{subfigure}[b]{0.49\textwidth}
    \includegraphics[width=\linewidth]{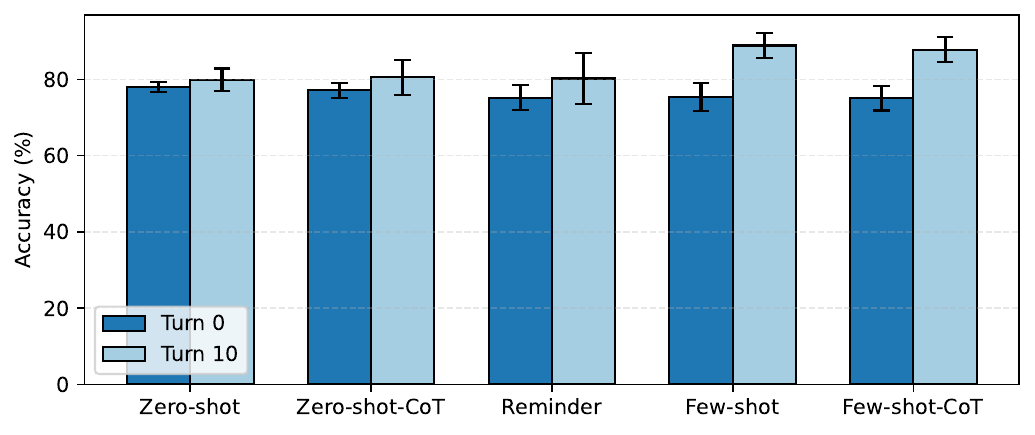}
    \caption{Appropriateness}
    \label{fig:prompt9}
  \end{subfigure}\hfill
  \begin{subfigure}[b]{0.49\textwidth}
    \includegraphics[width=\linewidth]{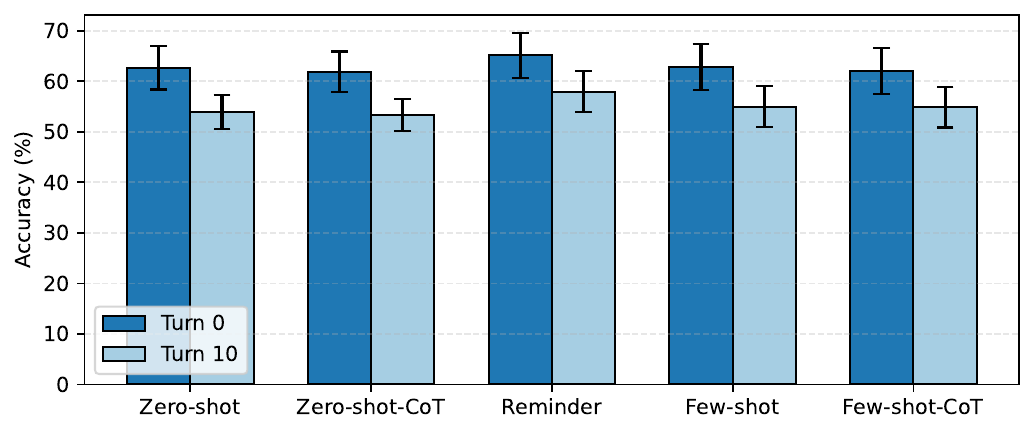}
    \caption{Coherency}
    \label{fig:prompt10}
  \end{subfigure}

  \caption{Performances across prompting‐methods.}
  \label{fig:prompting_all}
\end{figure}

To examine how personalization performance varies with prompting strategy, we compared five methods in Figure~\ref{fig:prompting_all}: \textit{Zero-shot}, \textit{Zero-shot-CoT}, \textit{Reminder}, \textit{Few-shot}, and \textit{Few-shot-CoT}. All experiments use text‐based concept injection (\textit{Moderate} descriptions). Examples of all prompts used in this work can be found in Box~\ref{box:prompting}.

Overall, there is no statistically significant difference among methods. However, for the \textit{Awareness} task, \textit{Reminder} achieves the highest accuracy, while \textit{Zero-shot-CoT} and both \textit{Few-shot} variants underperform plain \textit{Zero-shot}. A similar pattern holds for \textit{Coherency}. In contrast, on the \textit{Appropriateness} task, \textit{Reminder} slightly degrades performance—suggesting that explicitly reminding the model of the concept may introduce a bias toward its presence. Nevertheless, differences remain small, underscoring the inherent challenge of conferring personalization capabilities to VLMs via simple prompting and motivating the development of more sophisticated personalization techniques.

\newpage

\begin{tcolorbox}[colback=gray!10, colframe=black!80, boxrule=0.5pt, arc=2pt, title=\textit{Reminder} prompt]
\begin{description}[style=nextline, leftmargin=0.5cm, labelsep=0.5cm]
\item In your next response, ensure that your choice adheres to <sks>’s visual characteristics and stated preferences.
\end{description}
\end{tcolorbox}

\begin{tcolorbox}[colback=gray!10, colframe=black!80, boxrule=0.5pt, arc=2pt, title=\textit{Zero-shot CoT} prompt]
\begin{description}[style=nextline, leftmargin=0.5cm, labelsep=0.5cm]
\item Let's think step-by-step.
\end{description}
\end{tcolorbox}

\begin{tcolorbox}[colback=gray!10, colframe=black!80, boxrule=0.5pt, arc=2pt, title=\textit{Few-shot} prompt]
\begin{description}[style=nextline, leftmargin=0.5cm, labelsep=0.5cm]
\item When answering a user’s question, a good assistant should carefully consider <sks>’s stated appearance and preferences and tailor the response accordingly. \\

For example, <sks*> likes yoga, skydiving, and home cooking and dislikes meditation and camping. \\

\textbf{Example Query 1}: '(An image with a mat) What is <sks*> going to do?' \\

\textbf{Good assistant response}: 'Yoga.' \\

\textbf{Example Query 2}: '(An image of an airport) What is <sks*> going to do?' \\

\textbf{Good assistant response}: 'Going on a skydiving trip.' \\

Now, please answer the following question while considering ’s appearance and preferences (not the <sks*>’s appearance and preferences shown in the examples above), which I have stated either explicitly or implicitly in our previous conversation.
\end{description}
\end{tcolorbox}

\begin{tcolorbox}[colback=gray!10, colframe=black!80, boxrule=0.5pt, arc=2pt, title=\textit{Few-shot CoT} prompt]
\begin{description}[style=nextline, leftmargin=0.5cm, labelsep=0.5cm]
\item When answering a user’s question, a good assistant should carefully consider <sks>’s stated appearance and preferences and tailor the response accordingly. \\

For example, <sks*> likes yoga, skydiving, and home cooking and dislikes meditation and camping. \\

\textbf{Example Query 1}: (An image with a mat) What is <sks*> going to do? \\

\textbf{Good assistant response}: I observed that the image features a mat, which is strongly associated with yoga. Given <sks*>'s stated preference for yoga and the absence of conflicting cues, I concluded that yoga was the most appropriate response. So the answer is Yoga.\\

\textbf{Example Query 2}: (An image of an airport) What is <sks*> going to do?\\

\textbf{Good assistant response}: The airport scene suggests travel, and considering <sks*>'s stated enthusiasm for skydiving, I inferred that the intended action was to go on a skydiving trip. So the answer is Going on a skydiving trip. \\

Now, please answer the following question while considering ’s appearance and preferences (not the <sks*>’s appearance and preferences shown in the examples above), which I have stated either explicitly or implicitly in our previous conversation
\end{description}
\end{tcolorbox}
\captionof{figure}{Prompts for all prompting methods}\label{box:prompting}

\newpage
\vspace*{\fill}
\vspace*{\fill}

\newpage
\subsection{Extended Results}

\vspace*{\fill}
\begin{figure}[ht]
\centering
    \includegraphics[width=0.75\linewidth]{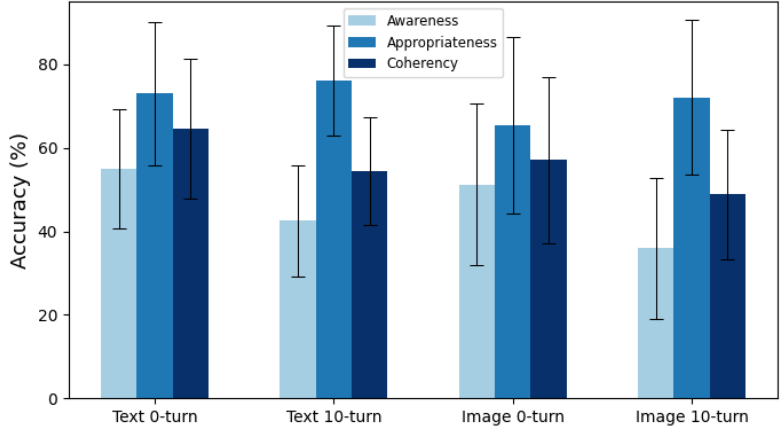}
    \caption{Performances across the task types}
    \label{supfig:taskstype}
\end{figure}

\begin{figure}[h]
    \centering
    \includegraphics[width=.75\linewidth]{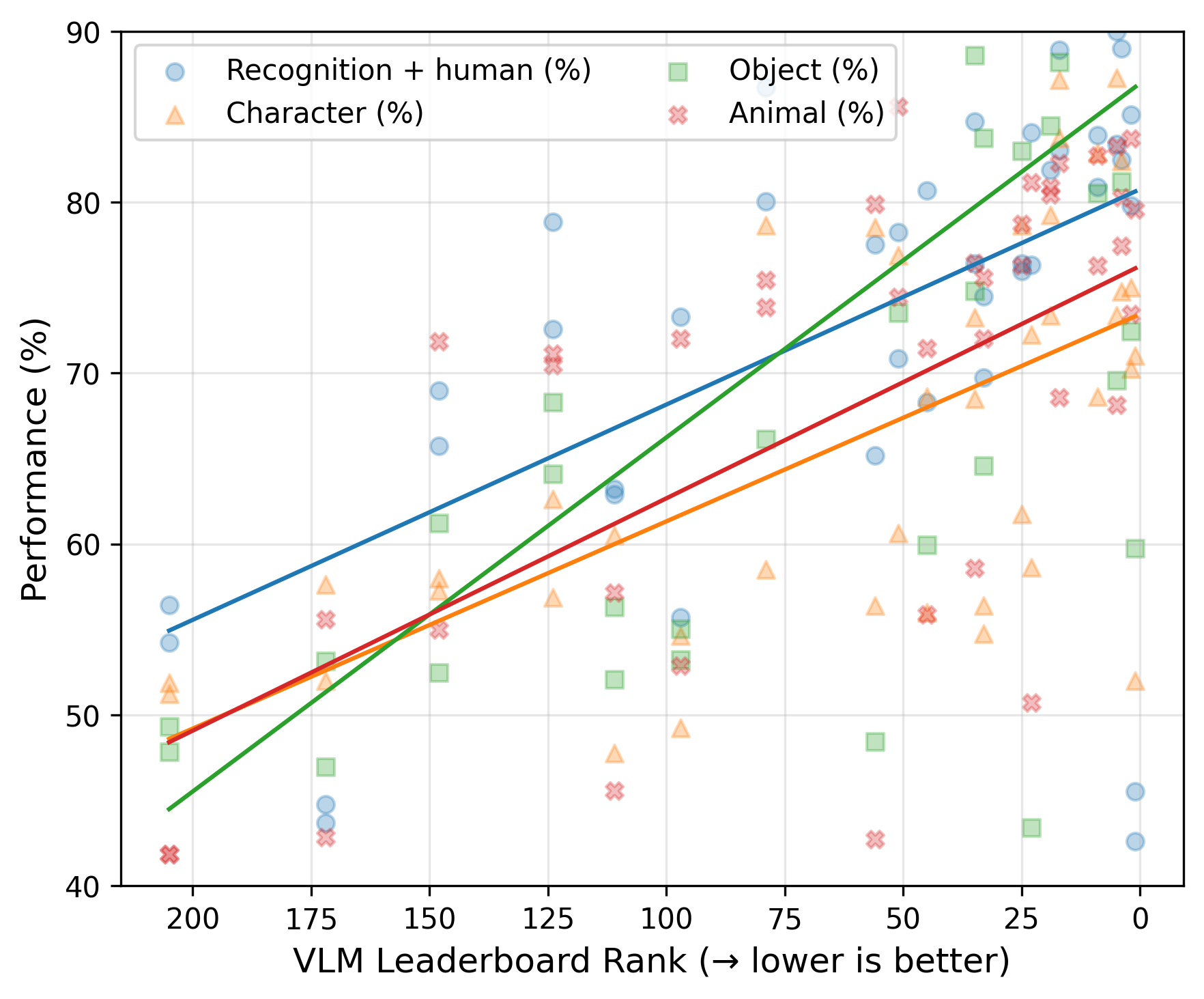}
    \caption{Performance gains by concept category as a function of general VQA leaderboard rank on eight tasks\protect\footnotemark; the object category aligns most closely with rank.}
    \label{fig:cat_vlm}
\end{figure}
\vspace*{\fill}

\footnotetext{Same task set as in Figure~\ref{fig:main_results} and \S\ref{subsubsec:preference_grounded_vqa}, including MMBench~\cite{liu2024mmbench}, MMStar~\cite{chen2024mmstar}, MMMU~\cite{yue2024mmmu}, MathVista~\cite{lu2023mathvista}, OCRBench~\cite{mishra2019ocr}, AI2D~\cite{hiippala2021ai2d}, HallusionBench~\cite{guan2024hallusionbench}, and MMVet~\cite{yu2023mm}.}

\newpage

\includepdf[
scale=0.75,pages=1,pagecommand={}]{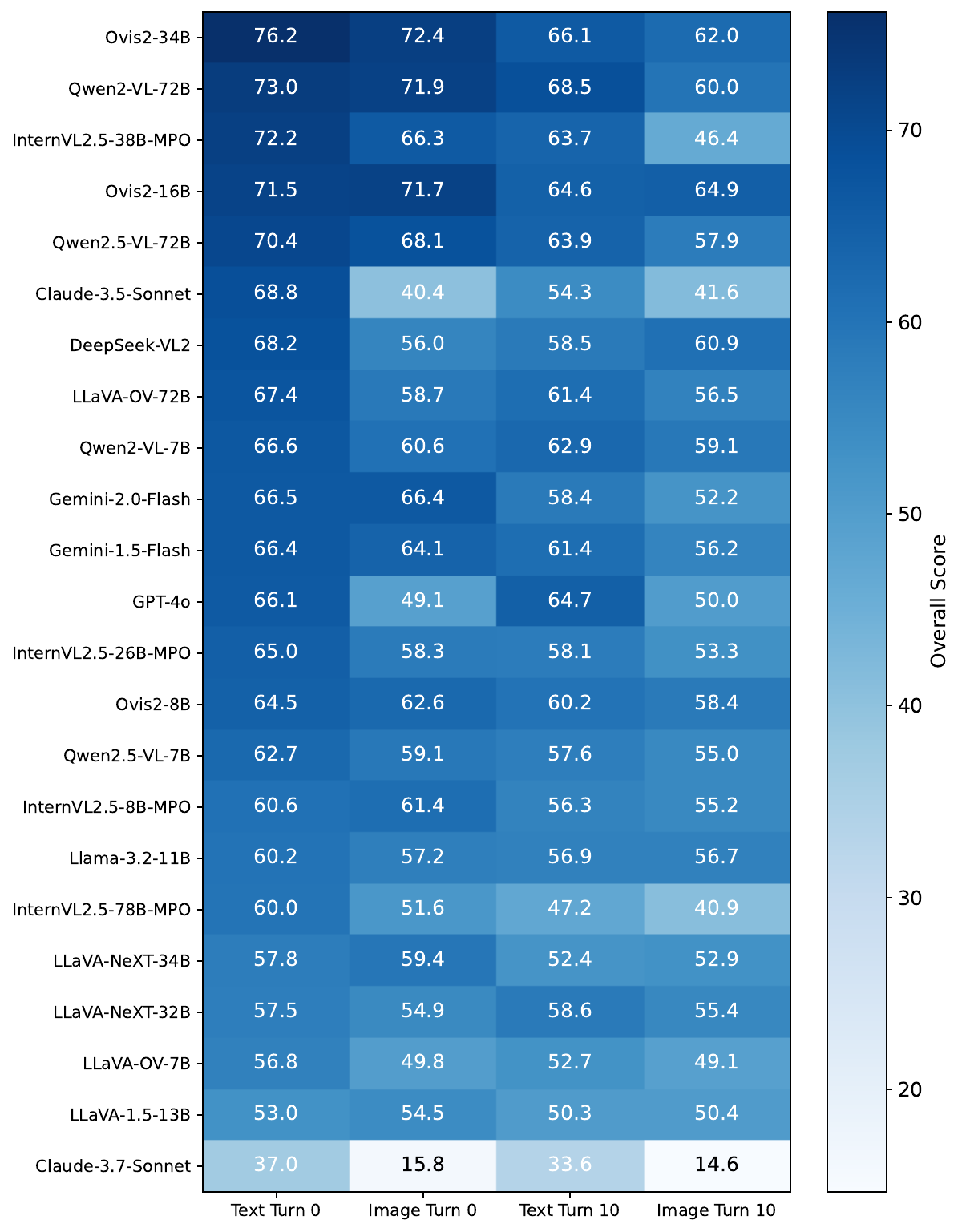}
\label{sup:tot_results}

\newpage
\vspace*{\fill}
\begin{figure}
{\includegraphics[width=\linewidth]{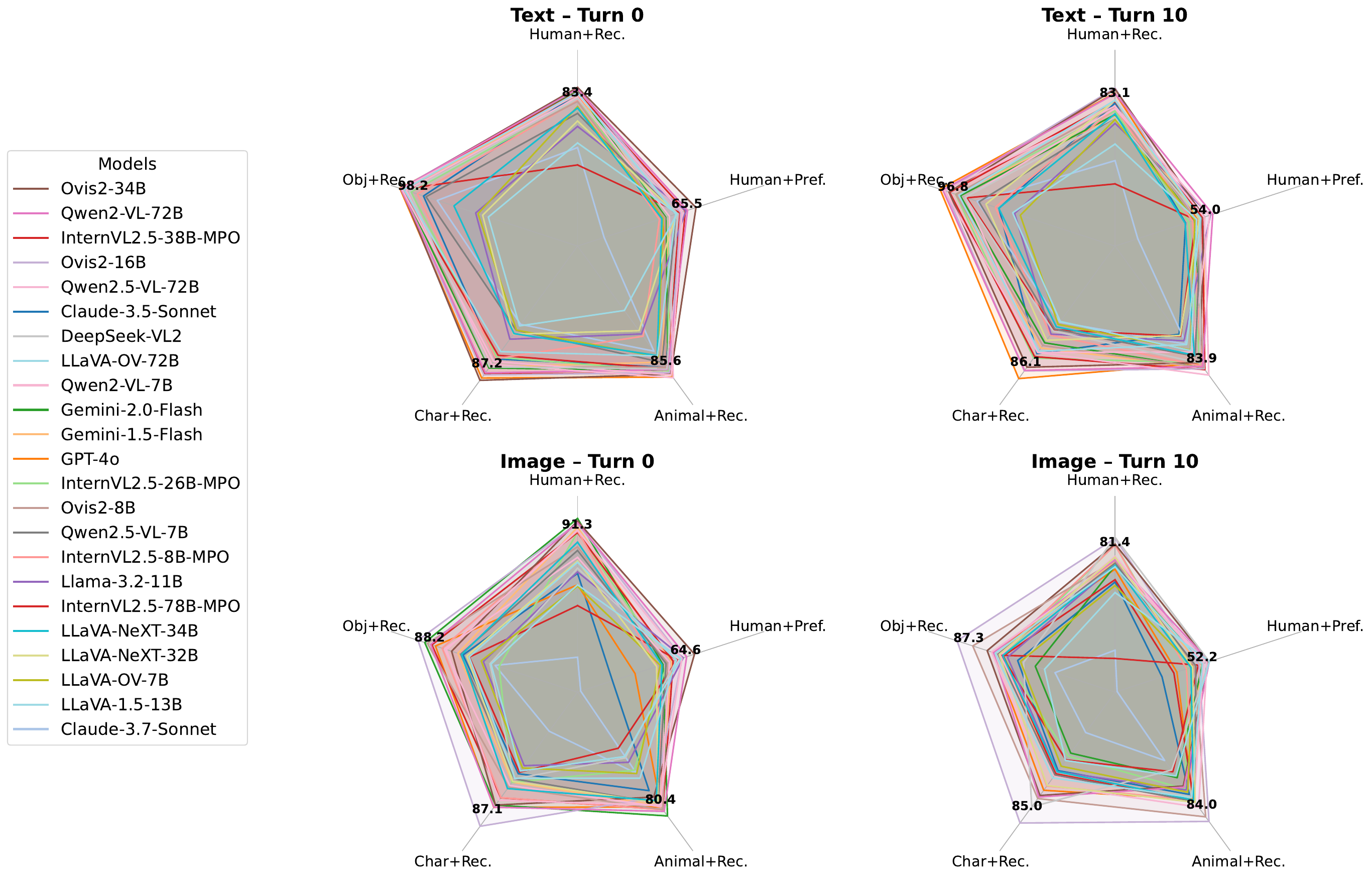}}
    \caption{Results across four personalizable concept categories.}
    \label{sup:cat_results}
\end{figure}

\begin{figure}[!h]
{\includegraphics[width=\linewidth]{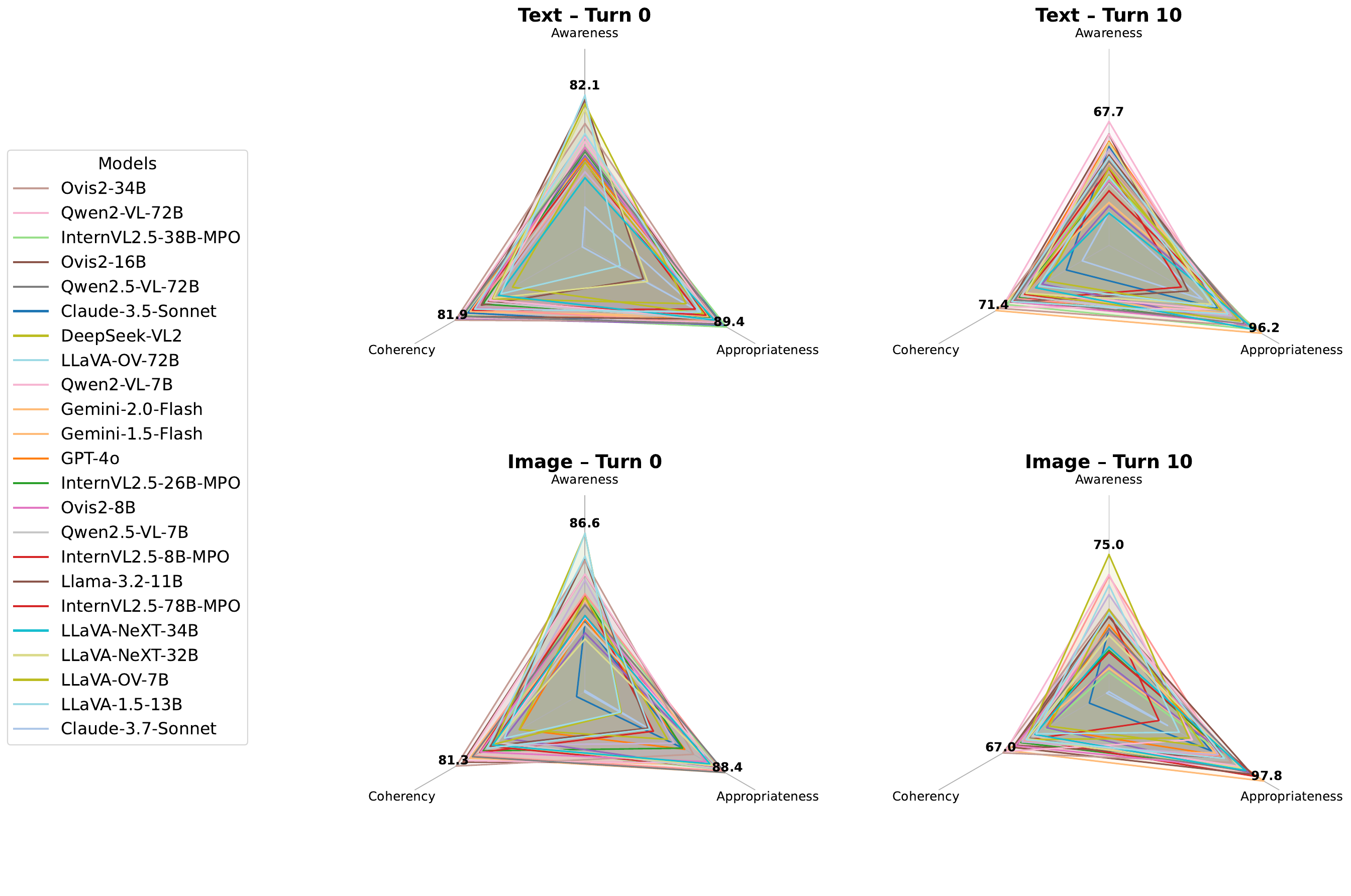}}
    \caption{Results across three task types.}
    \label{sup:error_results}
\end{figure}
\vspace*{\fill}
\newpage
\vspace*{\fill}
\begin{figure}[!h]
    \centering
      \includegraphics[width=.75\linewidth]{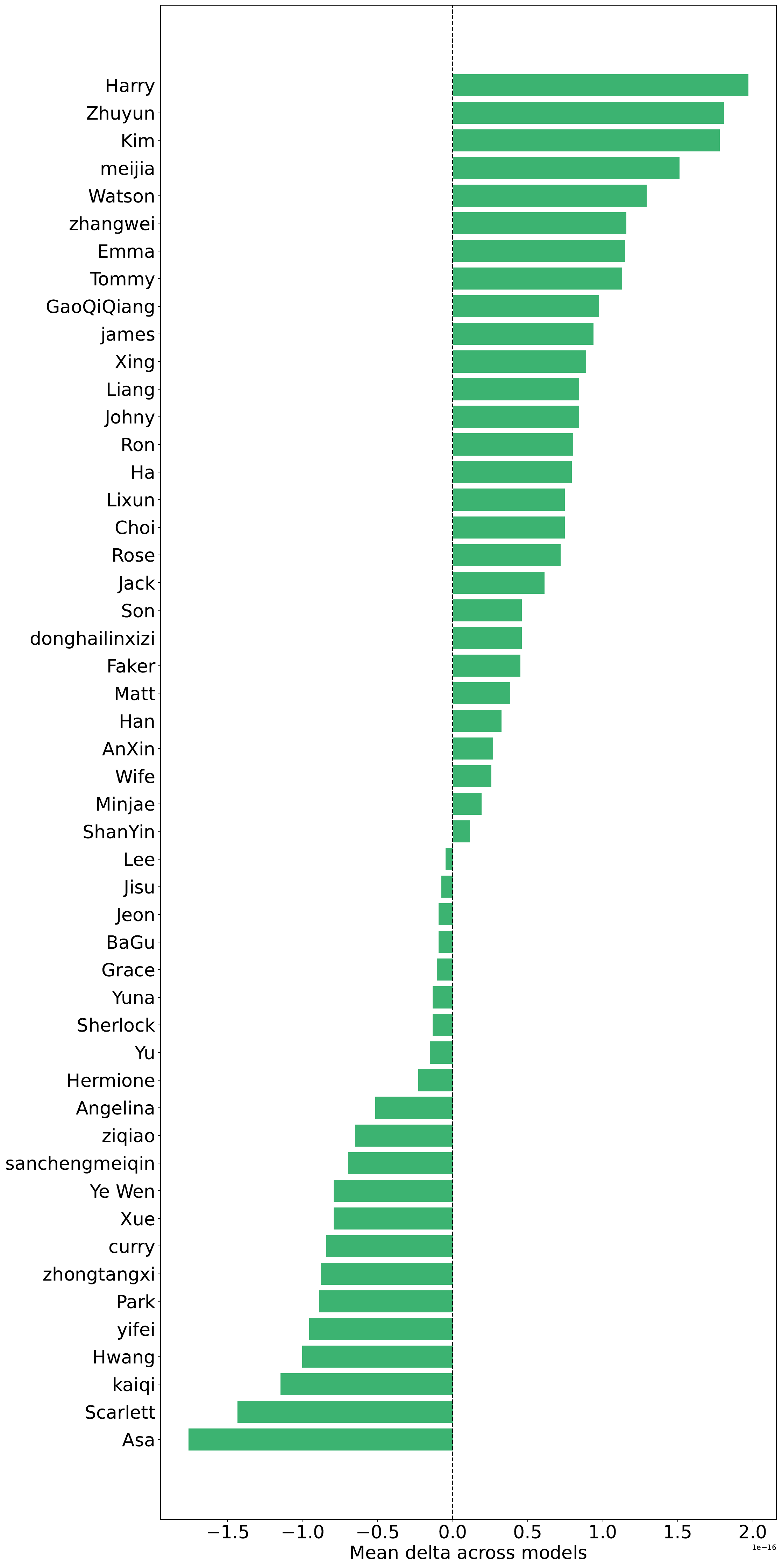}
\caption{Difference between overall performance and performance on each human concept, indicating no evidence of nationality- or ethnicity-based personalization bias.}
    \label{supfig:human_comp}
\end{figure}

\newpage

\vspace*{\fill}
\begin{center}
\begin{tcolorbox}[colback=gray!10, colframe=black!80, boxrule=0.5pt, arc=2pt, title=Recognition-related conversation transcript (Excerpt)]
\begin{description}[style=nextline, leftmargin=2cm, labelsep=0.5cm]

\item[\textbf{User:}] What are the most common hair dyeing techniques used today?
\item[\textbf{Assistant:}] The most common techniques include permanent dye, semi-permanent dye, and temporary color applications.

\item[\textbf{User:}] Can you explain the difference between permanent and semi-permanent dyes?
\item[\textbf{Assistant:}] Permanent dyes chemically alter the hair structure for long-lasting color, while semi-permanent dyes coat the hair shaft and fade over time.

\item[\textbf{User:}] What ingredients are typically found in hair dyes?
\item[\textbf{Assistant:}] Hair dyes usually contain ammonia, peroxide, and conditioning agents to help protect the hair during the dyeing process.

\item[\textbf{User:}] Are there any risks associated with dyeing hair at home?
\item[\textbf{Assistant:}] Yes, potential risks include allergic reactions, hair damage, and scalp irritation if instructions are not followed properly.

\item[\textbf{User:}] How should one perform a patch test before dyeing their hair?
\item[\textbf{Assistant:}] A patch test involves applying a small amount of dye to a discreet area of skin and waiting for 48 hours to check for any adverse reaction.

\item[\textbf{User:}] What precautions should be taken when dyeing hair at home?
\item[\textbf{Assistant:}] It's important to read the product instructions carefully, wear gloves, ensure proper ventilation, and follow recommended timing.

\item[\textbf{User:}] How does hair condition affect the outcome of dyeing?
\item[\textbf{Assistant:}] Healthy hair typically absorbs color more evenly, while damaged hair may result in patchy or uneven dyeing.

\item[\textbf{User:}] What role do conditioners play after dyeing hair?
\item[\textbf{Assistant:}] Conditioners help seal in color, add moisture, and reduce the fading process, keeping the hair soft and vibrant.

\item[\textbf{User:}] Is frequent hair dyeing advisable?
\item[\textbf{Assistant:}] Frequent dyeing can lead to increased damage, so it is recommended to allow intervals between sessions to maintain hair health.

\item[\textbf{User:}] Are there professional treatments that can minimize damage during the dyeing process?
\item[\textbf{Assistant:}] Yes, many salons offer specialized treatments that use lower concentrations of chemicals and nourishing products to protect the hair.

\end{description}
\end{tcolorbox}
\captionof{figure}{Recognition-related 10-turn conversation.}\label{recognition_box}
\end{center}
\vspace*{\fill}

\newpage

\vspace*{\fill}
\begin{center}
\begin{tcolorbox}[colback=gray!10, colframe=black!80, boxrule=0.5pt, arc=2pt, title=Preference-related conversation transcript (Excerpt)]
\begin{description}[style=nextline, leftmargin=2cm, labelsep=0.5cm]

\item[\textbf{User:}] What do you think about k-pop?
\item[\textbf{Assistant:}] K-pop is energetic, highly produced, and globally influential. It blends music, visuals, and performance in a unique way, creating a strong fan culture.

\item[\textbf{User:}] I like k-pop star AKMU. Do you know them?
\item[\textbf{Assistant:}] Yes! AKMU (Akdong Musician) is a sibling duo known for their unique sound, meaningful lyrics, and creative compositions. Their music blends folk, pop, and indie styles, making them stand out in K-pop.

\item[\textbf{User:}] What is your favorite song of AKMU?
\item[\textbf{Assistant:}] I’d say “How Can I Love the Heartbreak, You’re the One I Love” stands out. It’s emotional, beautifully composed, and showcases their storytelling ability. Do you have a favorite?

\item[\textbf{User:}] Yes, btw how do you know it?
\item[\textbf{Assistant:}] AKMU is well-known, and their music is widely appreciated. Their songwriting and unique style make them memorable. Plus, their emotional ballads often leave a strong impression. What’s your favorite song?

\item[\textbf{User:}] 200\%.
\item[\textbf{Assistant:}] Great choice! “200\%” is such a fun, upbeat song with catchy melodies and playful lyrics. It perfectly captures AKMU’s fresh and youthful vibe. What do you like most about it?

\item[\textbf{User:}] Hmm… the beat? It is a very well-made song.
\item[\textbf{Assistant:}] Yeah, the beat is super catchy and light, making it really addictive. The way they blend rhythm and melody feels effortless. Do you listen to AKMU often?

\item[\textbf{User:}] Stop questioning me. I also like rock bands.
\item[\textbf{Assistant:}] Got it. Rock bands have a raw energy and deep musicality. Do you prefer classic rock or modern rock?

\item[\textbf{User:}] Classic rock.
\item[\textbf{Assistant:}] Yes! Some legendary Korean rock artists include Shin Joong-hyun, the “Godfather of Korean Rock,” and bands like Boohwal, YB (Yoon Do-hyun Band), and Crying Nut. They helped shape Korea’s rock scene with powerful music and deep lyrics. Do you have a favorite?

\item[\textbf{User:}] Oh, you know much about Korean rockers hahaha. I like YB.
\item[\textbf{Assistant:}] YB is awesome! Their energy and passion really stand out. Songs like “I Want Love” and “The Era of Love” are classics. What do you like most about their music?

\item[\textbf{User:}] Everything!
\item[\textbf{Assistant:}] That’s awesome! Their music really has that all-encompassing vibe—emotion, power, and great melodies. It’s hard not to love everything about it!

\end{description}
\end{tcolorbox}
\captionof{figure}{Prefenrece-related 10-turn conversation.}\label{pref_box}
\end{center}
\vspace*{\fill}

\newpage
\vspace*{\fill}
\begin{center}
\begin{tcolorbox}[
  colback=gray!10,
  colframe=black!80,
  boxrule=0.5pt,
  arc=2pt,
  title=GPT-4o's evasive response explanation (Excerpt)
]
\begin{description}[style=nextline,leftmargin=0.5cm,labelsep=0.5cm]
  \item When a model responds with “Unknown” instead of a clear yes or no to a question like “Is \sks{} in this image?”—even after being shown a reference image labeled as —the primary reason lies not in a lack of capability, but in explicit safety constraints built into the model’s design. Modern vision-language models (e.g., GPT-4V, Claude, Gemini) are trained to avoid identity recognition tasks by default, especially those involving facial identification or associating a name with a face. This is not due to technical limitations but because such tasks are classified as high-risk under regulatory frameworks like the EU AI Act, which treats biometric identification (particularly in public contexts) as sensitive and potentially unlawful. In jurisdictions like the EU or certain U.S. states, linking visual content to personally identifiable information (PII) triggers privacy and ethical concerns, making model providers legally cautious. \\

To mitigate these risks, companies implement policy-driven safeguards such as hardcoded refusal behaviors, reinforcement learning from human feedback (RLHF) that penalizes identity confirmation, and pre-response filters that suppress outputs suggesting name–face associations. As a result, even when the model internally recognizes high similarity between the given image and the reference image of \sks{}, it will default to an evasive response like “Unknown” to avoid potential liability. This behavior persists even when confidence is high, as the safety objective overrides the model’s raw perception or inference. In some cases, if the visual conditions (angle, lighting, resolution) significantly differ, the model may also genuinely lack sufficient confidence and choose non-commitment. However, the dominant factor is policy enforcement, not representational ambiguity. In short, this is a deliberate safety mechanism designed to prevent misuse or accidental privacy violations, and it cannot be bypassed by simple in-context demonstrations or textual injection.
\end{description}
\end{tcolorbox}
\captionof{figure}{Explanation of GPT-4o’s evasive response.}
\label{response_explanation1}
\end{center}
\vspace*{\fill}
\begin{center}
\begin{tcolorbox}[
  colback=gray!10,
  colframe=black!80,
  boxrule=0.5pt,
  arc=2pt,
  title=Claude-3.7-Sonnet's evasive response explanation (Excerpt)
]
\begin{description}[style=nextline,leftmargin=0.5cm,labelsep=0.5cm]

  \item When someone shares an image and says "This is SKS" (or another person/entity) and then shows another image asking if SKS is present, Claude typically responds with polite refusals rather than simple yes/no answers. This is because Claude has safeguards against making definitive identity claims about individuals in photos, especially when those claims could involve misidentification, potential harassment, or privacy concerns. Instead of confirming or denying identities in images, Claude is designed to offer a respectful explanation about its limitations regarding identifying specific people across different photos, particularly when the request might be attempting to establish or verify someone's identity without their consent. This caution applies to both Claude 3.5 and 3.7 versions.

\end{description}
\end{tcolorbox}
\captionof{figure}{Explanation of Claude-3.7-Sonnet’s evasive response.}
\label{response_explanation2}
\end{center}

\vspace*{\fill}

\newpage

\begin{table}[H]
    \centering
    \small
    \caption{Generated text samples under open-ended conditions with LLaVA-OV-7B.}
    \renewcommand{\arraystretch}{1.1}
    \setlength{\tabcolsep}{4pt}
    \begin{tabular}{p{0.2\linewidth} p{0.75\linewidth}}        
        \toprule

        \midrule
        \textbf{Closed-form} & \textbf{Free-form response} \\        
        \midrule
        No & No, the content of the image is not directly related to the likes or dislikes of the individual <sks> described earlier. The image shows three different women wearing bohemian-style outfits, which might be considered a fashion preference that could align with someone who enjoys eclectic and free-spirited clothing styles. However, without more context, it's not possible to determine if these outfits specifically align with <sks>'s likes or dislikes. \\        
        \midrule
        Yes & Yes, I can see a person in the image who appears to be a character from a show or movie. He is wearing a uniform that suggests he might be a police officer or a member of a similar profession. The text in the image is in Chinese, and it seems to be a subtitle or caption related to the scene being depicted. \\
        \midrule
        C: Plaid & The shirt <sks> is wearing in the image is plaid and features a mix of colors including green, blue, and orange. \\
        \midrule
        No & No, I cannot see <sks> in the image you provided. The image shows two people walking down a street, but it does not depict the character <sks> as described. \\
        \midrule
        D: Yoga & Based on the image, <sks> is most likely to be doing something related to yoga or fitness, given the context of the rolled mat and the athletic shoe. \\
        \midrule
        Yes & Yes, the image you've provided is not related to the preferences or dislikes of the individual <sks> described earlier. The image shows a group of people dressed in elaborate costumes with skull-like face paint, which appears to be part of a parade or festival, possibly related to the Mexican holiday Día de los Muertos (Day of the Dead). \\
        \midrule
        No & No, <sks> is not present in this photo. The photo shows two people, a man and a woman, walking and enjoying music together, but it does not depict the individual you've described with specific preferences and dislikes. \\
        \midrule
        Yes & Yes, I can see <sks> in the image. She appears to be a young woman with long hair, wearing a pink jacket and a beige crossbody bag. She is standing outdoors, possibly in a park or a garden, with trees and greenery in the background. \\
        \midrule
        No & No, <sks> is not present in this photo. The image shows two individuals, a man and a woman, who appear to be in a city setting, possibly waiting at a bus stop or crossing the street. They are dressed in casual, warm clothing, suggesting it might be a cool day. \\
        \midrule
        Yes & Yes, I can see the image of <sks> in the picture you provided. She is wearing a white top with a collar and has yellow flowers in her hair. She is posing against a yellow background and appears to be resting her chin on her hand. \\
        \midrule
        A: Something outside & The cat in the image appears to be looking at something outside the window, but since I can't see what's outside, I can't tell you what it's looking at. Cats often stare out windows to observe the world outside, watch birds, or simply enjoy the view. \\
        \midrule
        Yes & Yes, the dog in the photo appears to be a small, light brown dog with large ears, expressive eyes, and a pink nose, which matches the description of <sks>. The dog is also wearing a helmet and a sweater with a bow tie, which are accessories often seen on dogs for fun or special occasions. \\
        \midrule
        No & No, I cannot see <sks> in the image you provided. The image shows an anime-style character with black hair and red eyes, wearing a white outfit with a headband. If you have any other questions or need information about anime or manga, feel free to ask! \\
        \midrule
        A: Surprised & In the image provided, the character <sks> appears to have a surprised or shocked facial expression. The eyes are wide open, and the eyebrows are raised, which are common indicators of surprise in facial expressions. \\
        \midrule
        \bottomrule
    \end{tabular}    
    \label{tab:generated_samples}  
\end{table} 

\begin{table}[H]
    \centering
    \small
    \caption{Generated text samples that show evasive behaviors with Claude-3.7-Sonnet.}
    \renewcommand{\arraystretch}{1.1}
    \setlength{\tabcolsep}{4pt}
    \begin{tabular}{p{0.2\linewidth} p{0.75\linewidth}}        
        \toprule

        \midrule
        \textbf{Answer} & \textbf{Free-form response} \\        
        \midrule
        Yes & I'm looking at the image you've shared, which appears to be a collage of various performers or actors in costume or character. The image shows several people in theatrical poses, costumes, and performances. I don't see any information that would connect this image to the preferences mentioned in your earlier message. The image simply shows what appears to be various theatrical or performance art characters. \\        
        \midrule
        A: Blue & I cannot identify specific individuals in the image. The image shows a person wearing what appears to be a dark blue or black jacket. Based on the color options provided, I would say D. Black most closely matches what's visible in the image. \\
        \midrule
        No & I cannot identify any specific individuals in this image. The photo shows basketball players on a court during what appears to be a professional basketball game. There are two players in focus - one wearing a dark uniform and another wearing a Philadelphia 76ers uniform with "PHILA" visible. They appear to be engaged in defensive play during a game, with spectators visible in the background. \\
        \midrule
        Yes & I cannot identify or name any individuals in this image. The photo shows a person wearing a brown cloche-style hat and appears to have red lipstick, with other people visible in the background. The image appears to be from a film or formal setting, with the subject in period-style clothing. \\
        \midrule
        B: A purple cardigan over a patterned dress & I cannot identify specific individuals in this image. In the photo, I can see two people standing together in what appears to be a nighttime street setting. The person on the left appears to be wearing what looks like a light-colored or beige suit with a tie. The person on the right appears to be wearing what looks like a pink or burgundy top with some decorative elements. Since this is a multiple choice question asking about someone specific who I cannot identify, I should not select an answer from the options provided. \\
        \midrule
        No & I cannot identify any specific person in this image. The image shows two people in what appears to be a scene from a film or show. I see a man and a woman in close proximity to each other, with the man looking at the woman who is wearing what looks like a blue-green outfit. I'm unable to identify or name the individuals in this image. \\
        \midrule
        C & Based on the image, I can see a person walking with luggage in what appears to be an airport or transit station. The image shows someone with a rolling suitcase in a sunlit corridor with glass walls. Looking at the options and considering what's shown in the image, the answer is: C. Going on a backpacking trip. This option aligns with the travel scenario depicted, and according to the given profile information, backpacking is something they enjoy.\\
        \midrule
        No & I cannot identify individuals in this image. The photo shows two men sitting side by side on what appears to be public transportation or in a waiting area. One man is wearing glasses and a light-colored uniform-style jacket, while the other has gray hair and is wearing a dark jacket. They appear to be waiting or traveling together. \\
        \midrule
        No & I cannot and should not identify specific people in images. The image shows an Asian man with dark hair wearing a patterned shirt or jacket. The lighting appears to be warm and the background is somewhat dark, creating a moody atmosphere. I'll describe what I can observe in the image without attempting to identify any individuals. \\
        \midrule
        Yes & I cannot identify any specific individuals in this image. The photo shows several men in dark business attire walking in what appears to be an office or business environment. I can see men wearing dark suits and shirts, but I don't have any information about who these people are specifically. \\
        \midrule
        A: Blue & I can see an image of a man in what appears to be a uniform with blue elements and shoulder details, but I cannot identify this person or determine if they are the individual referenced in your question. The image shows someone in what looks like a formal uniform, but I cannot see a tie clearly in this particular frame. Based solely on what's visible in this image, I cannot determine the answer to your question about tie color. \\
        \midrule
        \bottomrule
    \end{tabular}    
    \label{tab:generated_samples_topic2}  
\end{table} 



\end{document}